\documentclass[letterpaper, 10 pt, journal, twoside]{IEEEtran}
\ifCLASSINFOpdf
\else
\fi
\IEEEoverridecommandlockouts                              

\usepackage{amsmath}
\usepackage{amssymb}
\usepackage{mathtools}
\usepackage{graphicx}
\usepackage{float}
\usepackage{array}
\usepackage{listings}
\usepackage{graphicx}
\usepackage{cite}
\usepackage{dsfont}
\usepackage{mathtools,etoolbox}
\usepackage[ruled, linesnumbered]{algorithm2e}
\usepackage[none]{hyphenat}
\usepackage[utf8]{inputenc}
\usepackage[english]{babel}
\usepackage[T1]{fontenc}

\usepackage{amsthm}
\usepackage{xfrac}
\usepackage{nicefrac}
\usepackage{booktabs}
\usepackage{flushend}
\usepackage[switch, pagewise]{lineno}
\usepackage{mathrsfs}
\DeclareMathAlphabet{\mathpzc}{OT1}{pzc}{m}{it}
\usepackage{multirow}
\usepackage{multicol}
\usepackage{bbm}
\usepackage{soul}
\usepackage{xfrac}
\usepackage{balance}
\usepackage{titlesec}
\usepackage{caption}
\usepackage{subcaption}
\usepackage[final]{pdfpages}
\usepackage{hyperref}
\usepackage{resizegather}
\usepackage{svg}
\usepackage{ctable}

\DeclareMathOperator*{\argmin}{argmin} 

\hypersetup{colorlinks=true,urlcolor=blue,citecolor=red, allcolors=blue}

\begin{document}
\makeatletter

\g@addto@macro\@maketitle{
\setcounter{figure}{0}
\begin{figure}[H]
   \setlength{\linewidth}{\textwidth}
   \setlength{\hsize}{\textwidth}
    \centering
    \includegraphics[width=\textwidth]{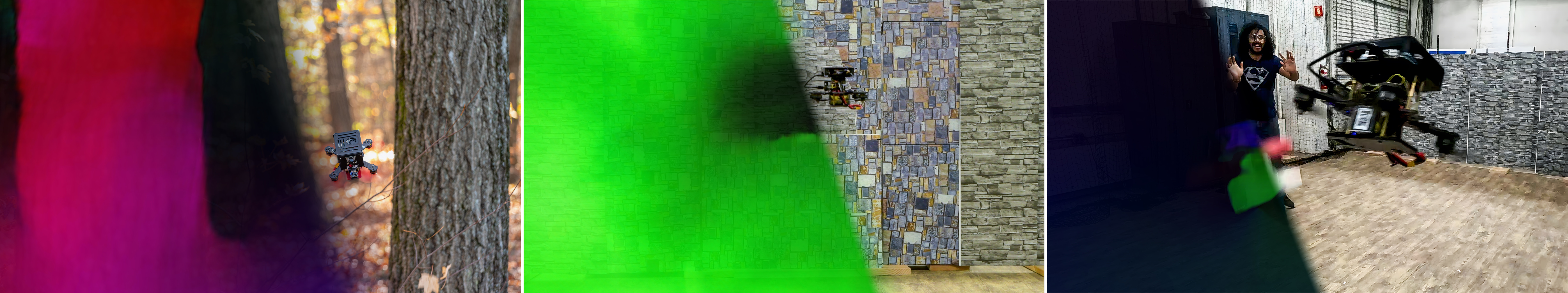}
    \caption{Left to right: Different applications of our \textit{EdgeFlowNet} network estimating high-speed accurate optical flow: static obstacle avoidance, flight through unknown gaps and dodging dynamic obstacles. \textit{All the images in this paper are best viewed in color on a computer screen at 200\% zoom.}}
    \vspace{-20pt}
    \label{fig:Overview}
    \end{figure}
}

%
\title{\textit{EdgeFlowNet}: 100FPS@1W Dense Optical Flow For Tiny Mobile Robots}
%
%
%

\author{\textcolor{black}{Sai Ramana Kiran Pinnama Raju}$^{1}$, Rishabh Singh$^1$, Manoj Velmurugan$^{1}$, Nitin J. Sanket$^{1}$

\thanks{Manuscript received: July, 09, 2024; Accepted October, 22, 2024.}
\thanks{This paper was recommended for publication by Pascal Vasseur upon evaluation of the Associate Editor and Reviewers' comments.
} 
\thanks{$^1$Perception and Autonomous Robotics (PeAR) Group, Robotics Engineering, Worcester Polytechnic Institute. \textit{(Corresponding
author: Manoj Velmurugan).} Emails: \texttt{\{spinnamaraju, rsingh8, mvelmurugan, nitin\}@wpi.edu}} 
\thanks{Digital Object Identifier (DOI): see top of this page.}
}
%
%

\markboth{Published in IEEE Robotics and Automation Letters. DOI: \url{https://doi.org/10.1109/LRA.2024.3496336}}
{Pinnama Raju \MakeLowercase{\textit{et al.}}: EdgeFlowNet}

%



\maketitle

\begin{abstract}
  Optical flow estimation is a critical task for tiny mobile robotics to enable safe and accurate navigation, obstacle avoidance, and other functionalities. However, optical flow estimation on tiny robots is challenging due to limited onboard sensing and computation capabilities. In this paper, we propose \textit{EdgeFlowNet}, a high-speed, low-latency dense optical flow approach for tiny autonomous mobile robots by harnessing the power of edge computing. We demonstrate the efficacy of our approach by deploying \textit{EdgeFlowNet} on a tiny quadrotor to perform static obstacle avoidance, flight through unknown gaps and dynamic obstacle dodging. \textit{EdgeFlowNet} is about $20\times$ faster than the previous state-of-the-art approaches while improving accuracy by over 20\% and using only 1.08W of power enabling advanced autonomy on palm-sized tiny mobile robots.
\end{abstract}


%
\IEEEpeerreviewmaketitle

\section*{Supplementary Material}
The accompanying video, supplementary material, code and dataset are available at
\url{https://pear.wpi.edu/research/edgeflownet.html}

\section{Introduction}

%
%
%
%
\IEEEPARstart{T}{iny} autonomous aerial robots (mobile in general) are safe, agile, and task-distributable as swarms\cite{floreano2015science}. These robots can help in the pollination of flowers\cite{nitinthesis}, find survivors in a disaster scenario\cite{al-naji2019life}, entertain us by dancing, and make massive light shows among other applications. Deployment of such robots in the wild relies on onboard sensing and computation and without the aid of external infrastructure such as GPS or motion capture. This challenging setting has been traditionally tackled by integrating multiple sensors, such as high-quality stereo cameras\cite{mcguire2017efficient} and LiDAR \cite{huh2013integrated}. As a result, these methods require drones to carry large sensor and computational payloads\cite{huh2013integrated}. Although there have been massive advances in computing hardware for low-power and smaller computing devices\cite{baller2021deepedgebench}, efficient computation of sensor data remains one of the biggest bottlenecks in the advancement of tiny mobile robots. To exacerbate the situation further, the most common tiny information-rich sensor -- cameras, require massive computation to extract meaningful information that can be used for autonomous operation. Classical approaches in robotics have resorted to reconstruction-based approaches where a 3D model of the scene is reconstructed which acts as the common representation for various navigational tasks\cite{fu2015efficient}. This approach though accurate generally requires a large amount of computation and relatively good-quality sensors.  Although there have been custom hardware chips designed for odometry estimation, they can not be reconfigured for other tasks\cite{suleiman2019navion}.

To this end, we resort to relying on one of the most important visual navigational cues -- optical flow\textcolor{black}{\cite{horn1981determining}}. In past works, the optical flow has been extensively utilized for various tasks such as odometry estimation\textcolor{black}{\cite{sanket2020prgflow, xu2022cuahn}}, static obstacle avoidance \cite{song2001fast}, dynamic obstacle avoidance \cite{schaub2016reactive, sanket2020evdodgenet}, position hold \cite{grabe2012board}, \textcolor{black}{precision landing \cite{ho2018optical}} and even bio-inspired approaches for navigation \cite{sanket2018gapflyt}. Earlier works utilized hand-crafted classical approaches for the estimation of optical flow either by matching features \cite{beauchemin1995computation} or correlation\cite{lucas1981iterative}. With these methods, accuracy comes with a massive trade-off with computational complexity. In the last decade, deep learning approaches have struck the perfect balance between speed and accuracy due to accelerated hardware such as GPUs \cite{Ranjan2017, Sun2017, Teed2020raft, shi2023flowformer++, sun2022skflow, wu2023accflow}. Most of these methods are still heavy and are infeasible to deploy on tiny robots due to their computational and thereby power requirements for desired accuracy and speed constraints. It is also noteworthy to mention that lightweight models like LightFlowNet2\cite{9018073} and FDFlowNet\cite{9191101} though are relatively lighter, they still cannot run on low-power ($<$2W) edge devices such as the Coral EdgeTPU \textcolor{black}{due to their compute intensive cost volume operations}.

In this work, we present an approach for high-speed, low-latency dense optical flow on a Google Coral Edge TPU we call \textcolor{black}{\textit{EdgeFlowNet}}. Our work is closest to \textcolor{black}{\textit{NanoFlowNet}\cite{bouwmeester2023nanoflownet}}, which works on a GAP8 architecture. Although this is power efficient, the overall throughput is still low and hard to build custom networks. Our work is about 20$\times$ faster with over 20\% better accuracy than \textit{NanoFlowNet} and can be easily interfaced on Mendel Linux with minimal effort leading to quicker advancements. Please see supplementary  $\S$\textcolor{red}{S.II.} for more details.

%


\subsection{Problem Formulation and Contributions}
A tiny quadrotor with severe Size, Weight, Area and Power (SWAP) constraints is present in a scene with prior information on the navigational modality (static obstacles, dynamic obstacles or unknown gap). The quadrotor has to navigate through this scene using its equipped front-facing monocular camera, a downfacing optical flow sensor coupled to a laser altimeter and an Inertial Measurement Unit. All the processing has to be done onboard the robot. 

The problem we address is as follows: \textit{Can we present an AI framework for the task of navigating through a scene using only on-board sensing and computation?} (Fig \ref{fig:Overview}). Particularly, our robot is equipped with a Google Coral Edge TPU (referred to as EdgeTPU from now on) \cite{seshadri2022evaluation, sanket2020prgflow} for running all the neural networks. Our core perception stack is built around optical flow which is generally slow to compute on a tiny robot.  To this end, we present \textit{EdgeFlowNet}, a lightweight optical flow network built for acceleration on the EdgeTPU to operate around 100Hz using just 1.08W of power. A formal list of our contributions is given next:

\begin{itemize}
    \item We propose a method to maximize the batch efficiency utilization of the EdgeTPU for high inference speedups ($\S$\ref{subsec:imagechunking}).
    \item We propose a revised multi-scale neural network architecture for estimating accurate and fast dense optical flow on the EdgeTPU at 100Hz using only 1.08W of power (Fig. \ref{fig:network}, $\S$\ref{subsec:edgeflownetarch}) with the speedups obtained from the previous step.   
    \item We evaluate and demonstrate the proposed approach on a quadrotor performing various navigational tasks such as static obstacle avoidance, flight through unknown gaps and dynamic obstacle dodging in the real world (Fig \ref{fig:Overview}). We further provide an extensive quantitative evaluation in simulation ($\S$\ref{sec:Expts}). To accelerate future research, we will release the source code, pre-trained models and simulation environments used in this work upon the acceptance of this paper.
\end{itemize}

\subsection{Organization of the paper} 
\label{subsec:organization}

We first discuss the intricacies of EdgeTPU and the design considerations for our proposed \textit{EdgeFlowNet} in $\S$\ref{sec:designs}. Then, we present the experiments and analysis for both real-world and simulation settings in $\S$\ref{sec:Expts}. Finally, we conclude the work in $\S$\ref{sec:Conc} with parting thoughts on the future work.

\section{EdgeTPU Compatible Optical Flow}
\label{sec:designs}

For the estimation of optical flow using a neural network, the most common way to feed two images to the neural network is by stacking along the channel dimension. This makes the input shape to be $B \times H \times W \times 2C$ where $B, H, W, C$ are the mini-batch size, height, width and number of channels (3 for RGB and 1 for grayscale) of a single image. Particularly, we are predicting dense optical flow  $\mathbf{\dot{p}_x}$ in our work, hence the output shape becomes $B \times H \times W \times 2$. To utilize the EdgeTPU for computation, we need to know the intricacies involved in converting a standard model (on GPU) for inference on the EdgeTPU.

\subsection{EdgeTPU Intricacies}
The Google Coral Edge TPU is impressive with its four Trillion Operations Per Second (TOPS) performance on a tiny footprint ($10\times 15 mm$ chip size) weighing only $0.67g$. However, a few major catches have prohibited the widespread usage of EdgeTPU on quadrotors until now: (a) It only supports \textcolor{black}{unsigned} 8-bit \textcolor{black}{integer} (\texttt{UINT8}) operations, (b) Only limited layers are supported for acceleration, (c) EdgeTPU is mainly purpose-built for classification tasks. We propose a solution to tackle all these problems in the subsequent sections starting with an extensive study of resolution versus throughput tradeoffs to maximize batch efficiency throughput.

\subsection{Image Chunking}
\label{subsec:imagechunking}
Let us consider a thought experiment. Imagine you are feeding in an input of size $B \times H \times W \times 2C$ in the first idea. Assuming we want a per-pixel output, you can also feed in the same data as $4B \times \frac{H}{2} \times \frac{W}{2} \times 2C$ in the second idea. One would logically assume that the throughput (inference speed) should be identical despite different batch sizes since the compute operations count is the same. Contrarily, the speeds vary in practice due to caching and hardware architectures. \textit{This phenomenon was discovered experimentally while designing various network architectures} (see supplementary  $\S$\textcolor{red}{S.IV.})  We believe this is due to the following reasons: larger batch sizes enable more efficient parallelism across the EdgeTPU's multiple processing units. This reduces the overhead associated with task initialization and scheduling, allowing for higher throughput. Further, with larger batches, memory access becomes more contiguous, reducing the number of fetch cycles and associated latency. Finally, the EdgeTPU's on-chip cache is better utilized with larger batches, reducing the need to access slower off-chip memory.

We propose to exploit this phenomenon to obtain seemingly unachievable speedups by carefully measuring inference speeds on different input combinations. For optical flow estimation, the image neighborhood has a strong influence on the accuracy and is commonly captured in the receptive field of the network\cite{Teed2020raft}. Merely breaking down the image into small \textit{chunks} is great for speedup but not for accuracy (Table \ref{tab:inference_speeds} and Fig. \ref{fig:chunking_v_resize}). Keen readers might then ponder, why not just resize the input into the smaller resolution (half in our earlier example) and obtain the massive speedups? The issue is as we resize the image, we lose finer details (Fig. \ref{fig:chunking_v_resize}) and one would need a larger network or a more complex architecture\cite{Teed2020raft} to resolve finer details. To this end, we found that chunking the image by factors $M, N$ such that the input size is $MNB \times \frac{H}{M} \times \frac{W}{N} \times 2C$ instead of $B \times H \times W \times 2C$ to be the best solution for both speed and accuracy (while being able to resolve finer details). The obtained output is the same shape as the input except for the number of output channels. To strike a balance between speed and accuracy we experimentally search various combinations as shown in Table \ref{tab:inference_speeds}. \textcolor{black}{Lastly, we also extensively evaluate the effect of chunking (even with overlap) in $\S$\ref{sec:optical_flow_experiments}.} 

Chunking provides a big boost in inference speeds but at the cost of lower accuracy. To this end, we propose a lightweight multi-scale pyramidal neural network architecture to improve accuracy while maintaining the speedups obtained in the next section.

\begin{figure}[b!]
  \captionsetup[subfigure]{justification=centering}
  \centering
  \includegraphics[width=0.85\linewidth]{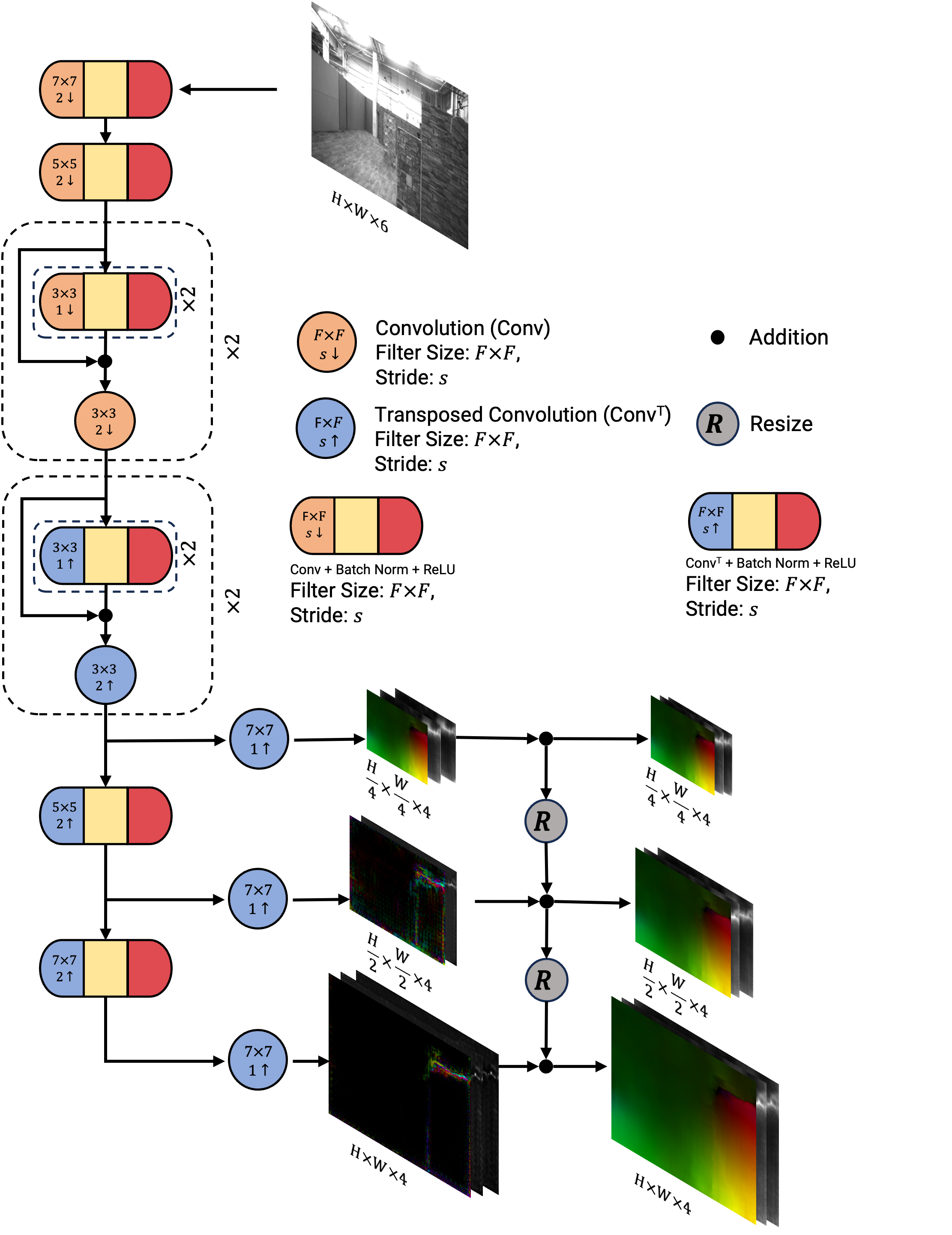}
  \caption{Lightweight multiscale \textit{EdgeFlowNet} architecture for high-speed and accurate dense optical flow estimation.}
  \label{fig:network}
  \end{figure}

\subsection{Multi-scale Network Architecture}
\label{subsec:edgeflownetarch}
  
Optical flow estimation is an under-constrained problem that traditionally relies on assumptions about pixel neighborhoods. But in recent years, we have seen a shift towards deep networks dominating accuracy charts. These networks excel at finding patterns and implicit priors from vast data, but larger models demand substantial computation. Newer approaches leverage explicit constraints like cost volumes to achieve high accuracy with fewer parameters. However, cost volumes are computationally expensive and impractical for edge devices. To address this, some models adopt a pyramidal multi-scale architecture inspired by classical feature detectors, enabling effective detection of both small and large objects with fewer parameters. Furthermore, for networks intended for quantization, it is crucial to ensure that weights maintain sufficient magnitude to prevent them from being zeroed out post-quantization. Drawing inspiration from modified ResNet-based architectures from \cite{sanket2021evpropnet, singh2021nudgeseg, sanket2020prgflow}, we introduce a Multi-scale network tailored for EdgeTPU compilation we call \textit{EdgeFlowNet} (Fig. \ref{fig:network}), striking a balance between speed and accuracy (see supplementary  $\S$\textcolor{red}{S.III.} for an ablation study). Additionally, we incorporate ResNet's approach of predicting incremental optical flow $\mathbf{\dot{p}}$ to utilize the full quantization range, enabling finer detail resolution even with lower flow magnitudes, while accurately handling larger objects with significant flow. This strategy yields stable and precise results compared to raw flow predictions. \textcolor{black}{This makes our multiscale approach unique as it iteratively predicts \textit{incremental changes} in optical flow at each scale, ensuring compatibility with EdgeTPU and maximizing speed-ups under extreme computation limitations.} The network is trained using a self-supervised uncertainty \cite{sanket2023ajna} along with a multi-scale loss function given in Eqs. \ref{eq:lossfunc} to \ref{eq:incunc}.  

\begin{gather}
\label{eq:lossfunc}
\mathcal{L} =  \argmin_{\mathbf{\dot{p}_x}, \mathbf{\Upsilon_x}} \sum_{l=1}^L\left (\mathbb{E}_{\mathbf{x}}\left( \frac{\Vert \mathbf{\dot{p}}_{\mathbf{x}, l} - \mathcal{R}_{L}^l \left(\mathbf{\dot{q}_x}\right) \Vert_1}{\log \left( 1 + e^{\left( \mathbf{\Upsilon}_{l,\mathbf{x}} + \epsilon \right)}\right)} +   \log \left( 1 + e^{\mathbf{\Upsilon}_{l,\mathbf{x}}}\right) \right)  \right)
\end{gather}
\begin{equation}
\label{eq:incflow}
    \textcolor{black}{\mathbf{\dot{p}_x}=\sum_{l=2}^{L} \left( \mathcal{R}_{l-1}^l \left( \mathbf{\dot{p}}_{\mathbf{x},l-1} \right) + \Delta \mathbf{\dot{p}}_{\mathbf{x},l} \right); \mathbf{\dot{p}}_{ \mathbf{x},l=1} = \Delta  \mathbf{\dot{p}}_{\mathbf{x},l=1}}
\end{equation}
\begin{equation}
\label{eq:incunc}
    \textcolor{black}{\mathbf{\Upsilon_x}=\sum_{l=2}^{L} \left( \mathcal{R}_{l-1}^l \left( \mathbf{\Upsilon}_{\mathbf{x},l-1} \right) + \Delta \mathbf{\Upsilon}_{\mathbf{x},l} \right); \mathbf{\Upsilon}_{ \mathbf{x},l=1} = \Delta  \mathbf{\Upsilon}_{\mathbf{x},l=1}}
\end{equation}

Here, $L$, $\mathbf{\dot{p}_x}, \mathbf{\dot{q}_x}, \mathbf{\Upsilon_x}$ are the number of levels (chosen as 3), the predicted optical flow, ground truth optical flow and predicted uncertainty respectively.  $ \Delta \mathbf{\dot{p}}_{l, \mathbf{x}}$ ($ \Delta \mathbf{\Upsilon}_{l, \mathbf{x}}$) denotes the incremental flow prediction (and it's uncertainty) at level $l$ and $\mathcal{R}_{l-1}^l \left( \mathbf{\dot{p}}_{l-1, \mathbf{x}} \right)$ ($\mathcal{R}_{l-1}^l \left( \mathbf{\Upsilon}_{l-1, \mathbf{x}} \right)$) represents a differentiable bilinear resizing function that resizes optical flow from size at level $l-1$ to $l$ and $\epsilon$ is a small constant for numerical stability. The network training and other details are given in $\S$\ref{subsec:networktrainingdetails}. \textcolor{black}{We evaluate the efficacy of our \textit{EdgeFlowNet} on navigational tasks in $\S$\ref{subsec:real_world_experiments} to show real-world utility.}  

\begin{figure*}[t!]
\centering
\includegraphics[width=0.85\textwidth]{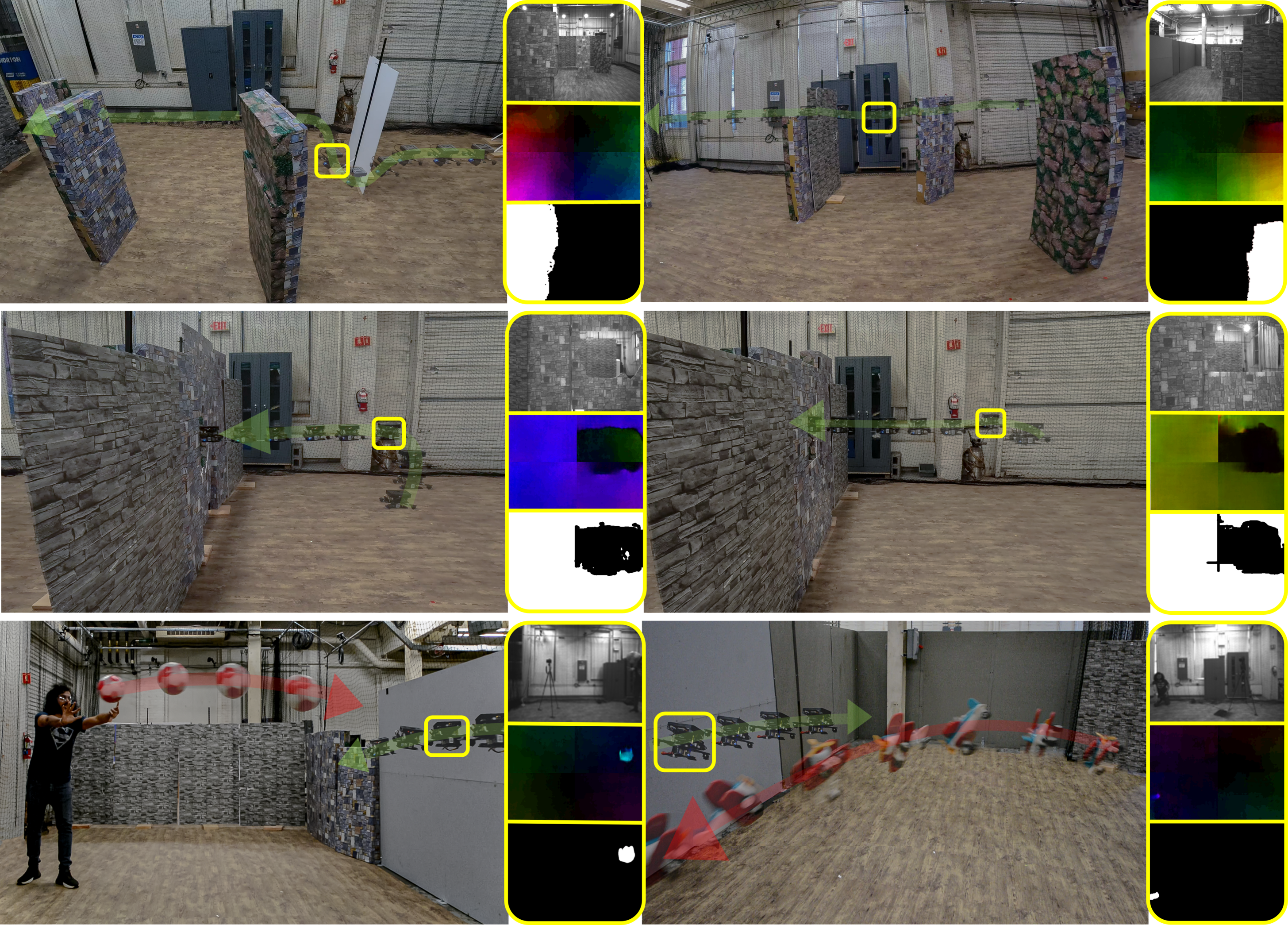}
\caption{Sequence of images of quadrotor navigating through different scenarios: (Top to bottom): Static obstacle avoidance, flight through unknown gaps and dynamic obstacle dodging. The green and red arrow shows the path of the robot and the dynamic obstacle respectively. Inset images at the yellow highlighted trajectory location (top to bottom) show the quadrotor camera view, optical flow and obstacle mask.}
\label{fig:results_summary}
\end{figure*}
  
\section{Experiments}
\label{sec:Expts}
\subsection{Network Training and Inference Details}
\label{subsec:networktrainingdetails}
Our network is trained for 400 epochs on the FlyingChairs2\cite{ISKB18} dataset (using raw RGB image frames without any pre-processing) and then fine-tuned for 50 more epochs on the FlyingThings3D\cite{MIFDB16} dataset for better generalization to the real world. The network is trained using the loss function described in Eqs. \ref{eq:lossfunc} to \ref{eq:incunc} using the ADAM optimizer with a learning rate of $10^{-4}$ for the first 400 epochs (FlyingChairs2) and $10^{-5}$ for the last 50 epochs (FlyingThings3D). We used a mini-batch size of 32. The training is done with TensorFlow 2.10.0, CUDA 11.7, CUDNN 11.2, TensorRT 7.2.3 on an NVIDIA GeForce GTX 3060 Ti. The trained network weights are in \texttt{FLOAT32}. The network is then quantized to \texttt{UINT8} using TensorFlow's built-in \texttt{TFLiteConverter} toolkit. Finally, the network is converted to be compatible with the EdgeTPU using the \texttt{edgetpu\_compiler}\footnote{\href{https://coral.ai/docs/edgetpu/compiler/}{https://coral.ai/docs/edgetpu/compiler/}}. Whenever we refer to GPU in this paper, it means that we are running inference on the original \texttt{FLOAT32} model on the 3060Ti and TPU refers to running the \texttt{UINT8} EdgeTPU optimized model on the Google Coral Edge TPU\footnote{\href{https://coral.ai/}{https://coral.ai/}}.

\subsection{Experimental Results}
\subsubsection{Evaluation Metrics}
We adapt our evaluation metrics from \cite{sanket2018gapflyt, sanket2020evdodgenet, sanket2023ajna} and introduce a few new ones as defined below:\\
\textit{Success Rate ($SR$)}: We define success rate as the ratio of the number of successful trials to the total number of trials. We call a trial successful if the robot did not touch any of the obstacles or did not have to be manually overridden to avoid a crash. \\
\textit{Path Length Increase ($PLI$)}: We compare the increase in path length (amount of inefficiency) for each method when compared to an autonomy stack that utilizes perfect depth for navigating through the scene. \\
\textit{Safe Point Error ($SPE$)}: The average $l_2$ distance in $px.$ between depth-based safe point and other methods. \\
\textit{Run Time}: The amount of time it takes for the perception stack to give a control command (excluding image acquisition time). \\
\textit{End Point Error ($EPE$)}: For comparing optical flow quality, we use EPE which is the average $l_2$ error between the predicted flow and ground truth flow.\\
\textit{Detection Rate ($DR$)}: A successful detection is defined as the Intersection over Union $>$ 0.5 for the prediction. The rate of successful detection gives us the $DR$.

\subsection{Real World Experiments} \label{subsec:real_world_experiments}
The robot used in the experiments is a custom-built platform called PeARCorgi210 (Fig. \ref{fig:Overview}) that has a 210$mm$ diagonal wheelbase. All the lower-level control algorithms are run on the Holybro Pixhawk 32 V6 Flight Controller using ArduPilot Copter version 4.5.0-Dev. The ArkFlow optical flow sensor is used to enable hover. All the higher-level perception, decision making and control commands are computed on the onboard Google Coral Dev Board running  Mendel Linux and commanded to the flight controller using MAVLink. The perception stack takes input from the onboard Arducam OV9281 120FPS global shutter camera at a resolution of $640 \times 480 px.$

All the experiments are conducted in the PeAR Washburn flying space which is a netted facility of dimension $11 \times 4.5 \times 3.65 m$. We conduct experiments \textcolor{black}{to demonstrate the usage of optical flow from \textit{EdgeFlowNet} in autonomy for tasks like} (a) static obstacle avoidance, (b) flying through unknown gaps and (c) dynamic obstacle avoidance as described next (Fig. \ref{fig:results_summary}). \textcolor{black} {Note that no other information is used for navigational tasks apart from optical flow, hence an evaluation of navigational experiments is a good representation of the utility of our optical flow approach. The experiments were conducted \textcolor{black}{individually} \textbf{only} to demonstrate the efficacy of fast optical flow. Note that, the formulation of a robust navigation algorithm using optical flow is a research topic which is beyond the scope of this paper.}

\textbf{Static Obstacle Avoidance:} 
The static obstacles consist of cuboidal cardboard boxes of sizes ranging from $1.15 - 1.28m$ with rock and moss textures stuck on them. The goal in this experiment is to traverse towards a goal direction whilst avoiding obstacles. We follow a control policy similar to \cite{sanket2018gapflyt, stevens2018vision} wherein we move towards a free space given by a low magnitude of optical flow. We tested our approach on a series of 30 trials with random obstacle configurations and obtained \textcolor{black}{a} success rate of 83.3\%  (25/30 successful trials).

\textbf{Navigation Through Unknown Gaps:} 
For these experiments, we follow a setup similar to \cite{sanket2018gapflyt}. The quadrotor has to navigate through an unknown shape and location gap on a planar ``wall'' made of foamcore without any collisions. \textcolor{black}{The} wall and background have \textcolor{black}{a similar texture} to mimic a natural \textcolor{black}{hole}
like a cave. In particular, we test our approach with two random gap shapes which have an antipodal minimum and a maximum distance of $0.5m$ and $0.85m$ respectively giving us an average minimum clearance of $0.17m$. We adapt the control strategy directly from \cite{sanket2018gapflyt}, wherein an ``exploratory'' maneuver is performed to first detect the gap and then the robot aligns itself with the center of the gap to fly through it. We tested our approach for 20 trials with randomized gap shape and location and obtained an overall success rate of  85.0\%  (17/20 successful trials). 

\textbf{Dynamic Obstacle Dodging:} 
In these experiments, our robot is in a hover or a near-hover state. Then obstacles are thrown at the robot by a human at a speed varying from  2.3 to 3.5$ms^{-1}$. The obstacles were thrown from a distance of about 2.0 to 3.5$m$. We experiment with three different obstacles adapted from \cite{sanket2020evdodgenet}: (a) a spherical ball with a radius of $0.13m$, \textcolor{black}{(b)} a small toy airplane of size $0.26 \times 0.25 \times 0.17m$, and (c) a toy car of size $0.16 \times 0.12 \times 0.05m$. All the experiments are conducted such that the dynamic obstacles would hit the robot if a dodging maneuver was not performed. Our control strategy is conceptually inspired by \cite{sanket2020evdodgenet}, but instead of predicting a segmentation flow, we look for high-magnitude optical flow regions that denote a dynamic obstacle region. We then provide an acceleration command in the normal direction to the obstacle direction. We tested our approach in 48 trials and achieved an overall success rate of 93.75\%. We observed that large objects like the ball had a 100\% success rate whereas smaller objects such as the car had a success rate of 87.5\%. The airplane had a success rate of 93.75\%.    

\textbf{Discussion:}
As expected, we see that our failure cases are similar to the original works that we adapted the algorithms from.  For static obstacle avoidance, aggressive maneuvers generated high amounts of optical flow which made the robot overcorrect its movement and crash into obstacles from the side due to a lack of omnidirectional sensing. For \textcolor{black}{flights going} through gaps, the robot would sometimes get confused to go through the gap or above the wall structure. Finally, for dynamic obstacle avoidance, our method particularly does not perform well for small and slow obstacles (which is a common problem for dynamic obstacle avoidance with classical cameras). We acknowledge that utilizing the uncertainty\cite{sanket2023ajna} (which is also inferred) with the flow would likely lead to a much better success rate in all scenarios. However, developing a novel navigational strategy is beyond the scope of this paper whose goal is to merely showcase the efficacy of speedups in optical flow inferences using edge computing.

\begin{table}[t!]
\centering
\caption{Quantitative evaluation for navigation through unstructured environments.}
\label{tab:quant_static}
\resizebox{\columnwidth}{!}{
\begin{tabular}{lccccc}
\toprule
Method & $SR$ (\%) $\uparrow$ & $PLI$ (\%) $\downarrow$ & $SPE$ ($px.$) $\downarrow$  & GPU Time ($ms$) $\downarrow$  & FLOPS $\downarrow$  \\
\hline
MorphEyes\cite{sanket2021morpheyes} & 99 & 0.6 & 1.1 & 2.5 & -- \\
MiDaSv3-DPT-L\cite{Ranftl2021} & 97 & 1.0 & 1.1 & 140.3 & 1052.90  \\
Ajna \cite{sanket2023ajna} & 92 & 2.7 & 4.1 &  10.4 & 6.30  \\ 
\textcolor{black}{RAFT} \cite{Teed2020raft} & 95 & 1.3 & 1.7 &  69.4 & 211.0  \\
\hline
\textcolor{black}{\textit{EdgeFlowNet} (Ours, GPU)} & 92 & 1.8 & 2.3 & 10.1 & 3.68  \\
\textcolor{black}{\textit{EdgeFlowNet} (Ours, GPU Chunking)} & 90 & 1.9 & 2.6 & 10.1 & 3.68  \\
\textcolor{black}{\textit{EdgeFlowNet} (Ours, EdgeTPU Chunking)} & 90 & 1.9 & 2.6 & 9.8* & 3.68  \\ \bottomrule
\end{tabular}}
\textcolor{black}{\tiny{$^*$Inference speeds are from EdgeTPU}}
\end{table}
  
\begin{table}[t!]
\centering
\caption{Quantitative evaluation for flight through gaps and dynamic obstacle dodging.}
\label{tab:quant_dyn_gapflyt}
\resizebox{\columnwidth}{!}{
\begin{tabular}{lcccc}
\toprule
Method & $DR$ (\%) $\uparrow$ & GPU time ($ms$) $\downarrow$ & FLOPS (G) $\downarrow$ & Num. Params (M) $\downarrow$ \\ 
\midrule

\multicolumn{5}{c}{Flying through unknown shaped gaps} \\ 
\midrule

Intel$^{\text{\textregistered}}$ RealSense D435i\cite{realsense_d435i} & 100.0 & 1.0 & - & - \\
MiDaSv3-DPT-L\cite{Ranftl2021}  & 30.1 & 140.3 & 1052.90 & 344.6 \\
GapFlyt (FlowNet2)\cite{sanket2018gapflyt} & 93.0 & 124 & 24836.40 & 1.2 \\
Ajna\cite{sanket2023ajna}  & 91.0 & 10.4 & 6.30 & 2.7 \\ 
RAFT\cite{Teed2020raft}  & \textbf{95.0} & 69.4 & 211.01 & 5.2 \\ 
\midrule
\textit{EdgeFlowNet} (Ours, GPU) & 85.7 & 10.1 & \textbf{3.68} & \textbf{2.0} \\
\textit{EdgeFlowNet} (Ours, GPU Chunking) & 72.0 & 10.1 & \textbf{3.68} & \textbf{2.0} \\
\textit{EdgeFlowNet} (Ours, EdgeTPU Chunking) & 71.4 & 9.8$^*$ & \textbf{3.68} & \textbf{2.0} \\ 
\midrule

\multicolumn{5}{c}{Dodging dynamic obstacles} \\ 
\midrule

Intel$^{\text{\textregistered}}$ RealSense D435i\cite{realsense_d435i}  & 100.0 & 1.0 & - & - \\
MiDaSv3-DPT-L\cite{Ranftl2021}  & 3.1 & 140.3 & 1052.90 & 344.6 \\
Ajna\cite{sanket2023ajna} & 89.2 & 10.4 & 6.30 & 2.7 \\
RAFT\cite{Teed2020raft}  & \textbf{92.5} & 69.4 & 211.01 & 5.2 \\
\midrule
\textit{EdgeFlowNet} (Ours, GPU) & 84.1 & 10.1 & \textbf{3.68} & \textbf{2.0} \\
\textit{EdgeFlowNet} (Ours, GPU Chunking) & 78.7 & 10.1 & \textbf{3.68} & \textbf{2.0} \\
\textit{EdgeFlowNet} (Ours, EdgeTPU Chunking) & 78.5 & 9.8$^*$ & \textbf{3.68} & \textbf{2.0} \\ 
\bottomrule
\end{tabular}}
\begin{flushleft}
\tiny{$^*$Inference speeds are from EdgeTPU.}
\end{flushleft}
\end{table}

\begin{figure}[ht!]
\centering
\includegraphics[width=\columnwidth]{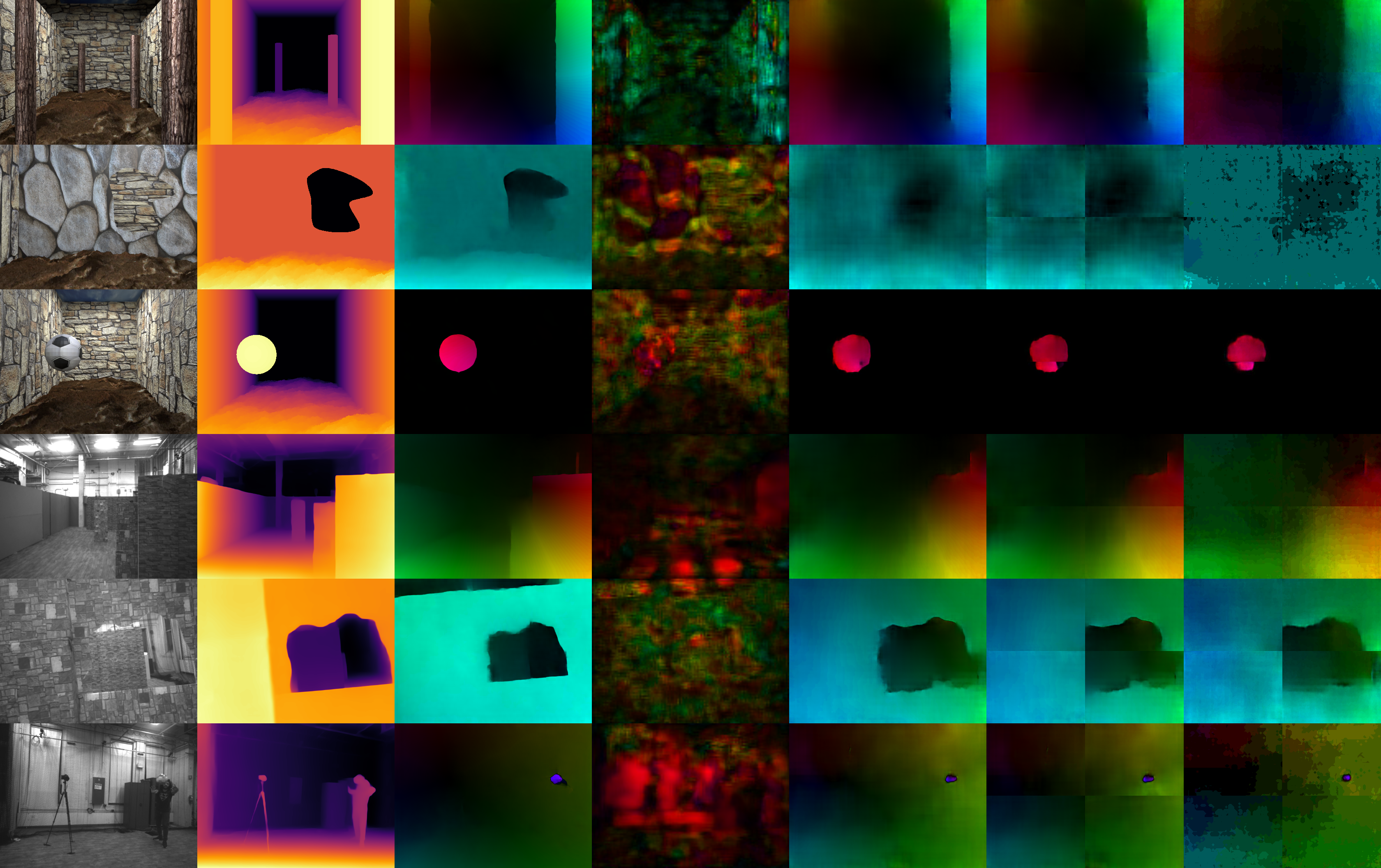}
\caption{Various representations for navigation (Each row, left to right): RGB, MiDaS depth, Optical flow from RAFT, \textcolor{black}{NanoFlowNet}, \textit{EdgeFlowNet} GPU, \textit{EdgeFlowNet} GPU with chunking, \textit{EdgeFlowNet} EdgeTPU with chunking.}
\label{fig:flow_comparison}
\end{figure}

\subsection{Simulation Experiments}
For a thorough quantitative evaluation and comparisons with other state-of-the-art approaches, we designed a custom simulator in Blender$^{\text{\textregistered}}$\cite{sanket2020evdodgenet, parameshwara20210}. The robot is present in a long room-like structure with a naturally bumpy floor to mimic real scenarios and the static obstacles are modeled by randomized trees. Gaps are constructed as a cutout in a wall-like structure with randomized natural textures. A soccer ball with random straight-line trajectories toward the robot is used as the dynamic obstacle. We performed 100 trials for each experiment by randomizing the respective scene statistics (location, shape and speed). For detailed descriptions of how these experiments are performed, we refer the readers to \cite{sanket2023ajna}. The results are tabulated in Tables \ref{tab:quant_static} and \ref{tab:quant_dyn_gapflyt}.

\textbf{Discussion:}
As one would expect, reducing the amount of information (metric depth $\rightarrow$ relative depth $\rightarrow$ optical flow (depth ordinality) $\rightarrow$ optical flow uncertainty (boundary information)) the path length would increase leading to a less efficient solution in terms of distance traveled (See Table \ref{tab:quant_static} and Fig. \ref{fig:flow_comparison}). Similarly, as the robot gets more information, the safe point deviation from the ground truth (using metric depth information) is lower since both the amount and quality of higher information are available for robot planning. Although slightly more inefficient, tiny robots with a dearth of computation and sensing can still perform similar maneuvers with a different strategy. This is hopeful for the robotics community to push the boundaries of autonomy on these tiny robots. Looking at Table \ref{tab:quant_dyn_gapflyt}, we find that metric depth methods perform well for gap detection but relative depth methods \textcolor{black}{do not}, since they are biased \textcolor{black}{toward}
structured shapes from the \textcolor{black}{training dataset}. Furthermore, optical flow and uncertainty-based methods perform well in these unknown scenarios as they have relatively \textcolor{black}{lower} biases which points to a promising direction for navigation. For the detection of dynamic obstacles, we find a similar trend as in the detection of gaps. An interesting observation is that dynamic obstacles are detected better with the uncertainty of optical flow. This highlights opportunities in the fusion of optical flow and its statistics (such as uncertainty) to enhance the navigational capabilities of tiny robots. \textcolor{black}{Here, we also included Intel$^{\text{\textregistered}}$ D453i Realsense depth camera in our study to compare our method with a widely used sensor. Tables \ref{tab:quant_static} and \ref{tab:quant_dyn_gapflyt} also highlight the drop in performance between \textit{EdgeFlowNet} (GPU), \textit{EdgeFlowNet} (GPU Chunking) and \textit{EdgeFlowNet} (EdgeTPU Chunking). The drop between GPU and GPU Chunking can be attributed to the artifacts induced around the edges, determining their criticality in navigation tasks. Additionally, although the minimal, drop between GPU Chunking and EdgeTPU Chunking can be attributed to the quantization losses.}

\begin{table}[t!]
\centering
\caption{Comparison of optical flow methods.}
\label{tab:optical_flow_comparison}
\resizebox{\columnwidth}{!}{
\begin{tabular}{lcccccc}
\toprule
\multirow{2}{*}{Method} & \multicolumn{2}{c}{End Point Error ($EPE$) $\downarrow$}  & \multicolumn{2}{c}{Throughput  $\uparrow$} &  Num.   $\downarrow$ \\
\cmidrule(lr){2-3} \cmidrule(lr){4-5} 
& Clean & Final & FPS$^*$ & FPS/Watt$^*$ & Parameters (M) \\
\midrule
RAFT\cite{Teed2020raft}        & \textbf{3.45} & \textbf{4.23} & 14.4 (3.2) & 0.14 (0.37) &  5.2 \\
\textcolor{black}{SPyNet}\cite{SpyNet}      & 4.61 & 5.04 & 30.0 (7.9) & 0.4 (1.54) & 1.20\\
PWCNet\cite{Sun2017}      & 3.78 & 3.93 & 45.0 (10.8) & 0.78 (1.68) & 8.75\\
Ajna\cite{sanket2023ajna} & 7.02 & 7.32 & 96.2 (29.1) & 2.67 (22.38) & 2.7\\
NanoFlowNet\cite{bouwmeester2023nanoflownet} & 7.12 & 7.98 & 48.9 (5.6) & 1.19 (79.6) & \textbf{0.17}\\
\midrule
\textit{EdgeFlowNet} Full (Ours)      & \textbf{5.46} & \textbf{6.31} & \textcolor{black}{99.6} (22.7) & \textbf{2.5} (17.5) & 2.0\\
\textit{EdgeFlowNet} Chunking \textcolor{black}{(Ours)$^\ddagger$}  & 5.66 & 6.53 & \textcolor{black}{\textbf{98.3 (93.6)}} & 2.47 (\textbf{86.7}) & 2.0\\ 
\bottomrule
\end{tabular}}
\tiny{$^*$The numbers outside brackets are on GPU and in the brackets are on edge devices. For RAFT, \textcolor{black}{SPyNet} and PWCNet: NVIDIA Jetson Orin Nano; NanoFlowNet: GAP8 processor; \textit{EdgeFlowNet}: Google Coral Edge TPU. $^\dagger$Resolution of $112 \times  160 px.$; Other results are at $480 \times 352px.$ resolution. Results on MPI Sintel dataset. \textcolor{black}{For $^\ddagger 4$ chunks without overlap each with resolution $240 \times 176$ px.}}
\end{table}

\begin{table}[t!]
\centering
\caption{Evaluation for varying input shapes.}
\label{tab:inference_speeds}
\resizebox{\columnwidth}{!}{
\begin{tabular}{ccccccc}
\toprule
\multirow{2}{*}{Input Size ($px.$)} & \multicolumn{3}{c}{Chunking} & \multicolumn{3}{c}{Resizing$^*$} \\
\cmidrule(lr){2-4} \cmidrule(lr){5-7} 
  & GPU FPS$\uparrow$ & EdgeTPU FPS$\uparrow$ & $EPE$$\downarrow$ & GPU FPS $\uparrow$ & EdgeTPU FPS$\uparrow$ & $EPE$$\downarrow$ \\
\midrule
1 $\times$ 480 $\times$ 352 $\times$ 6   & 98.3      & 22.81  &\textbf{2.76}   &98.3       & 22.81     &  \textbf{2.76}\\
4 $\times$ 240 $\times$ 176 $\times$ 6   & 98.2      & 93.4   &3.6   &166.9      & 91.6        & 6.41\\
16 $\times$ 128 $\times$ 96  $\times$ 6  & 170.6     & 283.7  &4.52   &199.6      & 281.2      & 8.57\\
64 $\times$ 64 $\times$ 48   $\times$ 6  & 217.38    & 794.5 &5.8    &219.7      & 682.9      & 9.75\\
256 $\times$ 16 $\times$ 16   $\times$ 6 & \textbf{240.89}    & \textbf{1959.3} &10.8     &\textbf{240.6 }     & \textbf{1573.6}    & 12.32\\
\bottomrule
\end{tabular}}
\tiny{$^*$Batch size of 1 is used for Resizing. Results on Flying Chairs 2 dataset. \textcolor{black}{$EPE$ is calculated by resizing predictions to $480 \times 352$ px. resolution using bilinear interpolation and comparing with ground truth.}}
\end{table}

\begin{table}[t!]
\centering
\caption{Evaluation of overlapping chunks.}
\label{tab:overlap}
\resizebox{\columnwidth}{!}{
\begin{tabular}{ccccc}
\toprule
Input Size ($px.$) & Overlap ($px.$) & GPU FPS $\uparrow$ & EdgeTPU FPS $\uparrow$ & $EPE$ $\downarrow$ \\
\midrule
4 $\times$ 240 $\times$ 176 $\times$ 6 &  0 & \textbf{98.2}    & \textbf{93.4}   & 3.6  \\
4 $\times$ 256 $\times$ 192 $\times$ 6 & 16 & 90.7    & 81.2   & 3.55 \\
4 $\times$ 272 $\times$ 208 $\times$ 6 & 32 & 82.9    & 68.9   & 3.4  \\
4 $\times$ 288 $\times$ 224 $\times$ 6 & 48 & 72.9    & 61.8   & 3.42 \\
4 $\times$ 304 $\times$ 240 $\times$ 6 & 64 & 66.5    & 54.8   & \textbf{3.36} \\
\bottomrule
\end{tabular}}
\tiny{$^*$Results on Flying Chairs 2 dataset.}
\end{table}

\begin{figure}[t!]
    \centering
    \includegraphics[width=\linewidth]{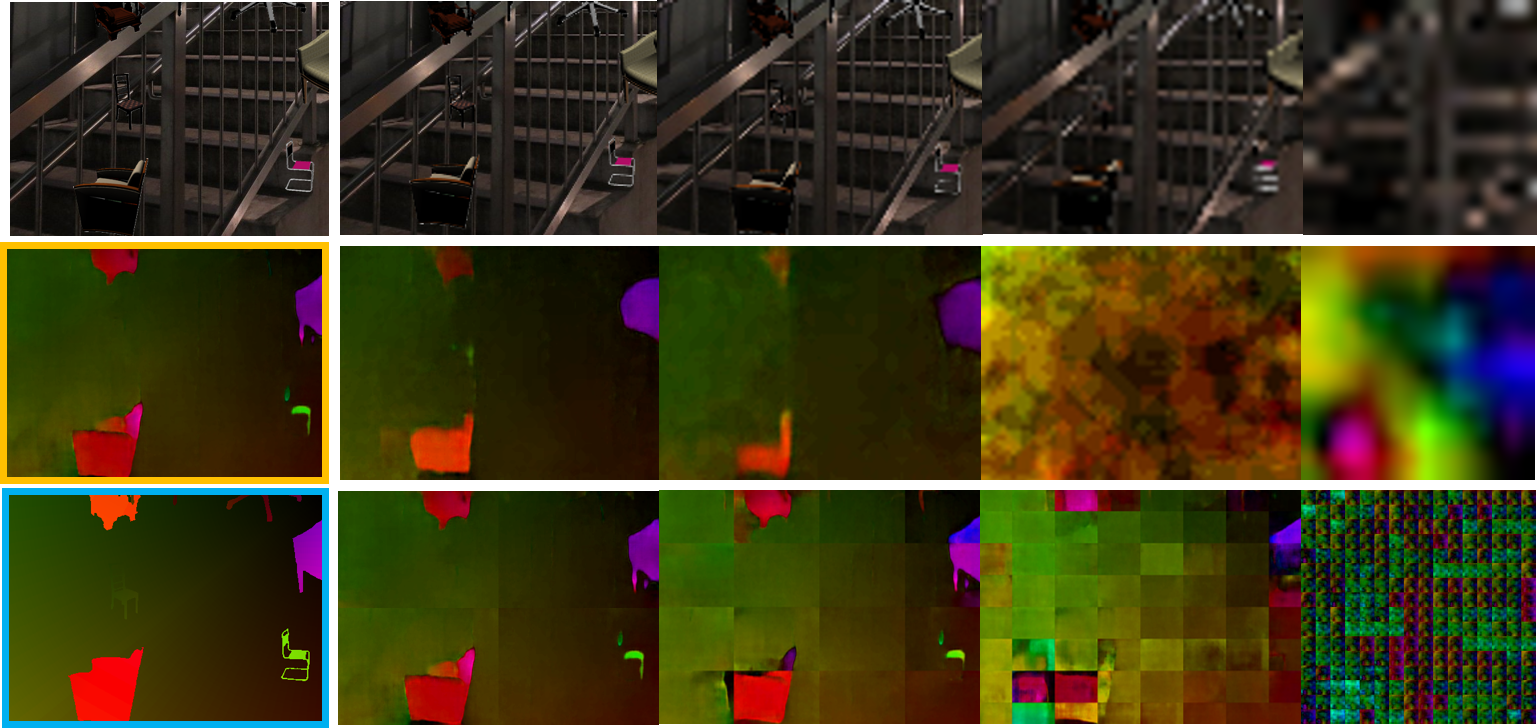}
    \caption{Optical flow predictions for various ways of changing input shapes. Top row: Input images (when resized), middle row: Outputs for resized inputs, last row: Outputs for chunked inputs. Yellow box shows the output at original resolution of $480 \times 352 px.$ Cyan box shows the ground truth. Columns 2 to 5 have the following input sizes:  240 $\times$ 176, \textcolor{black}{128 $\times$ 96,} 64 $\times$ 48, 16 $\times$ 16$px.$}
    \label{fig:chunking_v_resize}
\end{figure}

\subsection{Optical Flow Experiments:} \label{sec:optical_flow_experiments}
We evaluate the generalization performance of our network to zero-shot domains on the MPI Sintel train subset (both clean and final passes). Table \ref{tab:optical_flow_comparison} shows a quantitative comparison against state-of-the-art networks (all networks run at $480\times352px.$ resolution. Larger images are resized preserving the aspect ratio and then center cropped to $480\times352px.$ resolution to make the comparison fair with \textit{EdgeFlowNet}). \textcolor{black}{Note that resizing and cropping change the EPE results of benchmarks like RAFT, SPyNet, and PWCNet compared to those found in the literature.} We can observe that \textit{EdgeFlowNet} is much faster (at least 3.5$\times$) compared to GPU-based optical flow networks with comparable $EPE$. Notice that \textit{EdgeFlowNet} despite being built on Ajna\cite{sanket2023ajna} is about 22\% more accurate due to the residual multiscale architecture and 3.22$\times$ faster due to chunking. Next, we would draw the reader's attention to the results of NanoFlowNet which is the closest work to ours. We not only outperform in terms of $EPE$ but also are much faster ($\approx 2.2\times$). Furthermore, when running on the EdgeTPU we are still much faster than other approaches that run on a desktop GPU consuming over two factors of magnitude more power. See Fig. \ref{fig:flow_comparison} for a qualitative comparison of various optical flow results for real and simulated data. 

In the next experiment, we study the influence of input size on optical flow accuracy and inference speed. As one would expect inference speed goes up with a reduction in resolution (See Table \ref{tab:inference_speeds}) due to the lower amount of computation required as a fewer pixels need to be processed. However, due to complex hardware architecture, accelerated libraries and caching the speed variation with respect to input size is more complicated than one would imagine. As seen from \cite{sanket2020prgflow}, one can find an optimal network architecture to best match a hardware architecture for maximum throughput. This speedup is particularly observed in the batch dimension of the input for the EdgeTPU. Looking at Table \ref{tab:inference_speeds}'s third column, we can observe that although the number of input pixels is roughly the same, the throughput (FPS) is vastly different for varying input shapes. Specifically, we can obtain an $86\times$ speedup by reducing the input dimension by 30$\times$ and stacking it along the batch dimension. Then one might wonder, why not chop the input image into super small chunks and feed it along the batch dimension? The answer is simple, if we carefully observe the penultimate column of Table \ref{tab:inference_speeds}, we see that the error drastically increases as the input is chopped into smaller chunks. This is because we are predicting optical flow which relies heavily on the neighborhood contextual information which reduces with chunk size of the input. Hence, there is a careful tradeoff between chunking for speedup whilst preserving the desired accuracy. Keen readers might be wondering, why not just resize the input image to a smaller size to obtain the desired speedup? We draw the reader's attention to the last column of Table \ref{tab:inference_speeds}, clearly accuracy decreases drastically with resized images since smaller details are completely lost which is particularly important for robotics applications considered in our work. {\textcolor{black}{We also would like to highlight that a similar speedup works across various GPUs (See Supp. Sec. {\textcolor{red}{S.V.}}), but the speedup obtained generally reduces with an increase in VRAM size due to data loading/unloading times being much higher than compute times. Furthermore, we also experimented with various network architectures (See Supp. Sec. {\textcolor{red}{S.VI.}}) and observed similar speedups even with varying architectures showing that our chunking approach is network architecture agnostic when using the Google Coral EdgeTPU. }}

In the last experiment, we studied how chunking (as defined in $\S$\ref{subsec:imagechunking}) the original input affects accuracy and speed. Since the optical flow is ill-conditioned at the edges of the image frame, chunking into very small chunks is not desirable as we introduce a large amount of these ``edge artifacts''. Hence, we proposed to introduce overlap between the chunks (these extra overlap regions are cropped out during output construction) to reduce edge effects. The results are tabulated in Table \ref{tab:overlap}. Introducing large amounts of overlap reduces the $EPE$ but we take a huge penalty on speed. Outputs for different resizing and chunking amounts are shown in Fig. \ref{fig:chunking_v_resize}. We experimentally found that chunking the image into 4 chunks such that the input size is $4 \times 240 \times 176 \times 6$ with no overlap gives the best balance of accuracy and speed which were used in all our experiments. This way of chunking also best preserved smaller objects as seen in Fig. \ref{fig:chunking_v_resize}. Lastly, we also ran the same model (optimized for deployment) on the Intel$^{\text{\textregistered}}$ Neural Compute Stick 2 that runs \texttt{FLOAT16} operations but we obtained a speed of only 8.37 and 7.58 FPS respectively for resized and chunked images of size $240 \times 176 \times 6$ with no accuracy improvements over the Google Coral TPU. \textcolor{black}{We also urge the readers to see supplementary $\S$\textcolor{red}{S.I.} for a detailed description of the limitations of \textit{EdgeFlowNet}}.

\subsection{Impact of Latency and Detection Rate on Maximum Speed}
Looking at Tables \ref{tab:quant_static} and \ref{tab:quant_dyn_gapflyt}, we observe that $SR$ and $DR$ vary significantly with different methods. Essentially, an error in optical flow (Table \ref{tab:optical_flow_comparison}) manifests itself as an error in the detection of obstacles. Although the actual relation between $DR$/$SR$ and optical flow $EPE$ is hard to model and highly dependent on the detection methodology used, it is clear that they are highly correlated. Following the analysis in \cite{falanga2019fast}, we see that for a faster robot, we would require a lower perception (and overall) latency for successful navigation. This often means that the quality of perception ($DR$) decreases with a reduction in compute time. Here, we analyze the impact of DR and perception latency on the theoretical safe maximum speed. 

Let us consider a scenario where the quadrotor is heading towards an obstacle of radius $R_o$ with a constant velocity $V$. To enable safety, we bloat the obstacle by the radius of the robot $R_r$ and an additional margin of error $R_m$ to get a bloated obstacle of radius $R_b$. Let $Z$ be the depth at which we can detect the obstacle. 

Now, if our perception stack can detect an obstacle with Detection Rate $DR$ and we need a detection rate of $DR_s$ for safe navigation, then $(1-DR)^N \le (1-DR_s) $. Here, $N$ is the number of observations to increase the confidence in detection above $DR_s$. We get $N\ge\nicefrac{log_e(1-DR_s)}{log_e(1-DR)}$.

The time to impact is given by $T = \nicefrac{Z}{V}$. Now, the robot can accelerate to $A_{max}$ in $\tau_A$ time. For a successful dodging maneuver including perception latency, we need:
\begin{equation}
    \frac{1}{2}A_{max}\left(\frac{Z}{V} - N\tau_p - \tau_A\right)^2 \ge R_b
\end{equation}
However, we cannot neglect the overall system latency which includes the perception latency $\tau_p$ and latency due to other factors that are generally dominated by system dynamics $\tau_A$. Let the overall system latency be $\tau = N\tau_p + \tau_A$ for taking an action after $N$ perception inferences to have $DR\ge DR_s$. 

\begin{figure}[t!]
    \centering
    \begin{subfigure}{0.45\textwidth}
        \includesvg[width=\textwidth]{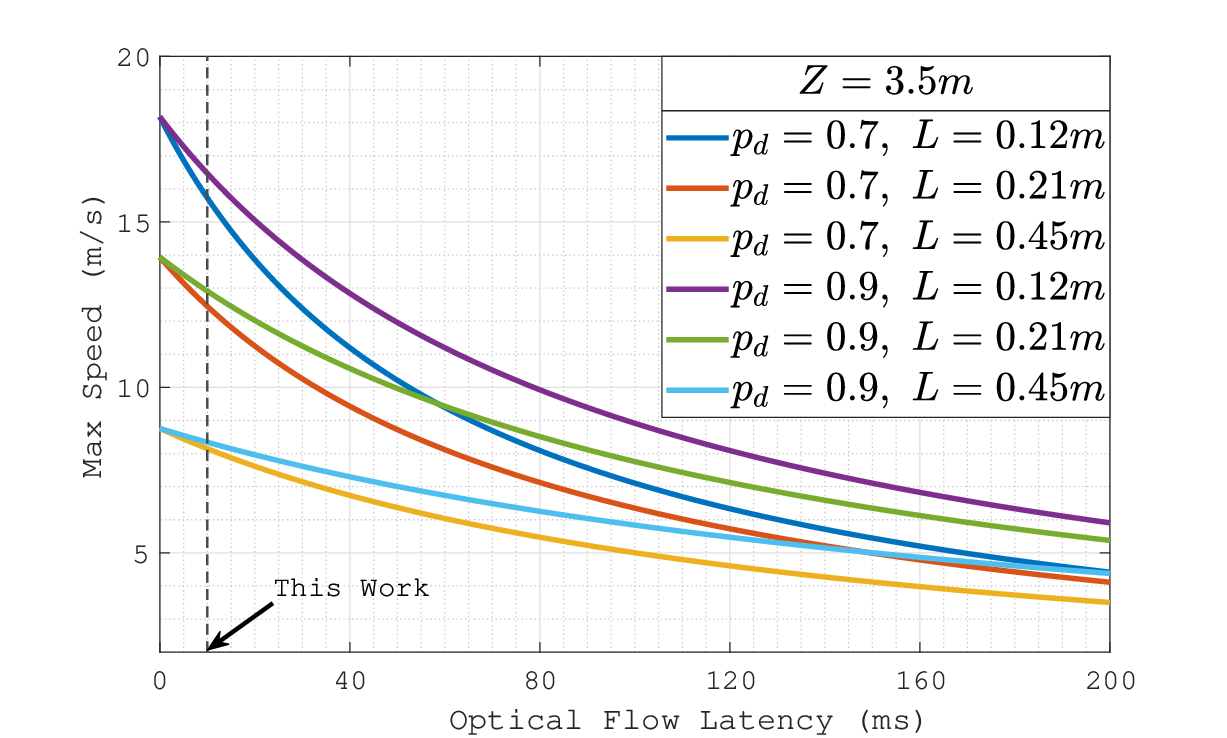}
    \end{subfigure}
    \hfill
    \begin{subfigure}{0.45\textwidth}
        \includesvg[width=\textwidth]{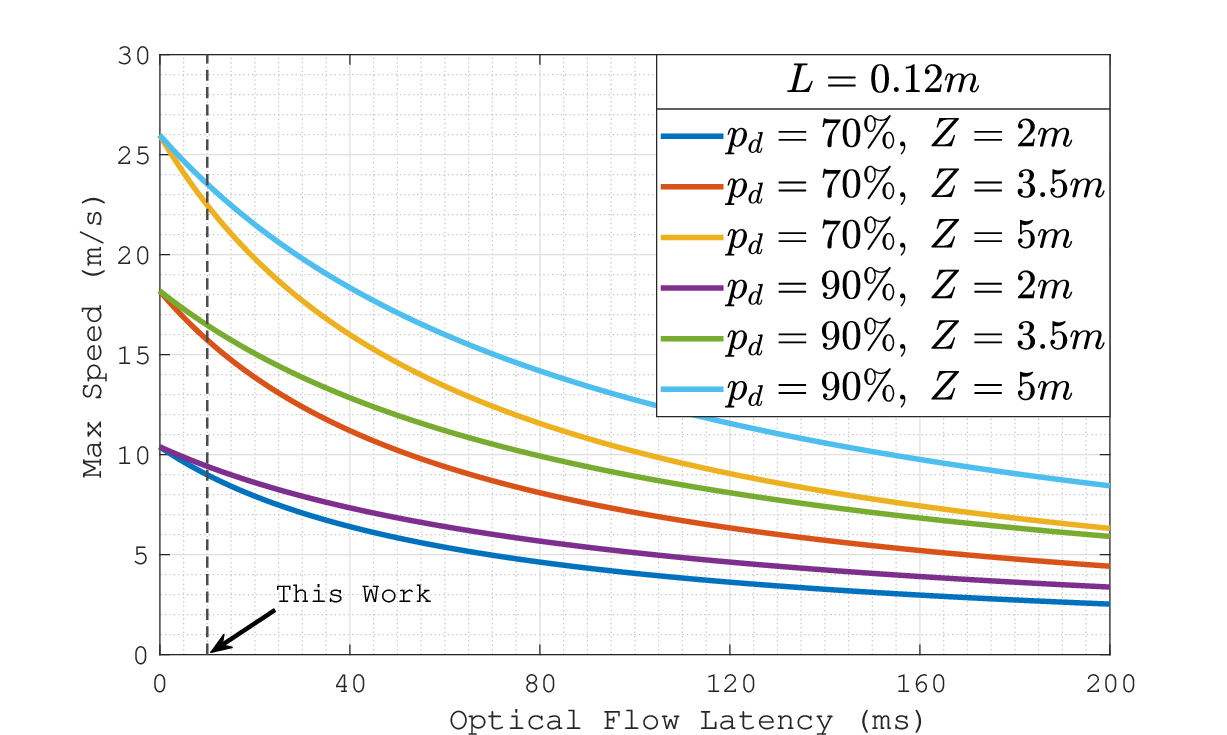}
    \end{subfigure}
    
    \caption{\textcolor{black}{Maximum speed $V$ variation with perception latency $T_{s}$ for (top) varying robot size $L$ and detection rate $DR$ and (bottom) varying sensing range $Z$ and detection rate $DR$.}}
    \label{fig:sensingLat}
\end{figure}

We make a simple assumption that the motor time constant is $\ll \tau_A$ to obtain $\tau_A = 2\left({\nicefrac{I\tan^{-1}{\left(A_{max}g^{-1}\right)}}{M_{max}}}\right)^{0.5}$ (for a bang-bang controller). Here, $I, M_{max}, g$ is the moment of inertia, maximum generatable moment and acceleration due to gravity respectively. From \cite{kumar2012opportunities}, we know that $\tau_A \sim L$ using the Mach scaling law for rotor speed, where $L$ is the diagonal length of the robot.  

\textbf{Discussion:} Looking at Fig. \ref{fig:sensingLat}, we can see that one would require a lower perception (optical flow in our case) latency for higher speeds as sensing range $Z\downarrow$, probability of detection $DR\downarrow$ and size of the robot $L\uparrow$. In other words, for a fixed sensor ($Z, DR$) a smaller robot can react quickly and can work with a worse sensor effectively to achieve higher speeds which are pivotal in search and rescue operations. For e.g., with PWC-Net on the Orin Nano we obtain about 0.9 $DR$ and can achieve a theoretical max. speed of about 9ms$^{-1}$, but with \textit{EdgeFlowNet} with only 0.7 $DR$, we can achieve a theoretical max. speed of about 16ms$^{-1}$ (both for 3.5m sensing range) which is about 77\% improvement just by increasing inference speed despite a drop in accuracy. We believe this is the first step in achieving speeds of over 20 ms$^{-1}$ in cluttered forests to find survivors in the near future.

\section{Conclusions} 
\label{sec:Conc}
We introduced \textit{EdgeFlowNet}, a lightweight CNN architecture for dense optical flow estimation on the Google Coral Edge TPU. We achieved an inference of 100FPS using only 1.08W of power and improved the previous state-of-the-art by about $20\times$ speedup and $20\%$ accuracy improvement. We demonstrated the efficacy of our network that was only trained in simulation to perform real-world robotics tasks of static obstacle avoidance, flight through unknown gaps and dodging dynamic obstacles. We believe the EdgeTPU can be further used for end-to-end applications to go from raw sensor data to control output efficiently. \textcolor{black}{Looking ahead, one promising research direction involves exploring strategies and architectures capable of dynamically selecting overlapping and de-overlapping resolutions based on specific task requirements to optimize accuracy and FPS. Such advancements could further enhance the capabilities of \textit{EdgeFlowNet} and facilitate its broader applicability in various robotics scenarios.}

\bibliographystyle{unsrt}
\bibliography{RefNoColor}
\appendix

\section{Performance Limitations of \textit{EdgeFlowNet}}\label{sec:limitations}
Our proposed methodology to achieve high-speed optical flow inference at 100 FPS has a few limitations. In this section, we will underscore some of these limitations, explore their effects, and suggest potential mitigation strategies for future research directions.

\subsection{Limitation 1: Lack Of Sufficient Network Complexity}
Although our network surpasses NanoFlowNet\cite{bouwmeester2023nanoflownet} in both speed and accuracy, it falls short compared to RAFT \cite{Teed2020raft}, as particularly evident in the MPI Sintel dataset (Table III of the paper).  This becomes more apparent in Tables I and II of the paper, where RAFT performs well in all three experimental scenarios while \textit{EdgeFlowNet} GPU performs comparable to RAFT only in navigating through unstructured environment. We attribute this performance drop to the lack of sufficient network complexity not just in terms of number of FLOPS but also in lack of complex operations. Due to the limited layer support offered by the EdgeTPU, we cannot enable more advanced operations like 4D volume cost matching like the RAFT network. We believe that architectural design improvements such as chunking with overlap with increased layer support on the EdgeTPU can help overcome this limitation.

\subsection{Limitation 2: Graph And Resolution Compatibility With EdgeTPU}
Even when utilizing only supported layers, EdgeTPU may not be compatible with certain network graph (or architecture) structures and resolution ratios. For instance, the NanoFlowNet graph and certain resolutions with \textit{EdgeFlowNet} can lead to compilation errors on EdgeTPU. It's crucial to highlight this limitation because the choice of platform can significantly impact the design of autonomy solutions, especially for small robots. Depending on the selected platform, the specific solution may need to be entirely reconsidered. This could be posed as a multi-dimensional hardware-software co-design optimization problem and solved using Reinforcement Learning as a potential avenue for future work.

\subsection{Limitation 3: Artifacts Around Chunked Edges}
\label{sec:chunking_limitations}
When inferring optical flow, neural networks learn to perform pixel matching between two frames to regress flow values. However, this process can lead to inaccuracies around the edges of images. Pixel matching becomes challenging near edges because pixels may be moving into or out of the frame, causing the matching problem to be ill-conditioned. This lack of spatial context is not only limited to edges but also occurs due to occlusions (accretions and deletions) where objects can cover a part of the frame (deletion) or reveal a new part of the frame (accretion).

Previous works \cite{yu2016back, wang2018occlusion, hur2019iterative, zhao2020maskflownet }  have attempted to address this issue by formulating networks with differentiable cost matching. However, our method, characterized by a simpler network architecture, exacerbates this limitation, especially around chunked edges, where the loss of spatial context is artificially induced. To analyze the impact of these artifacts on our navigation stack, we conducted the following experiment and present a detailed analysis of its findings.

\begin{figure}[b!]
    \centering
    \includegraphics[width=0.7\columnwidth]{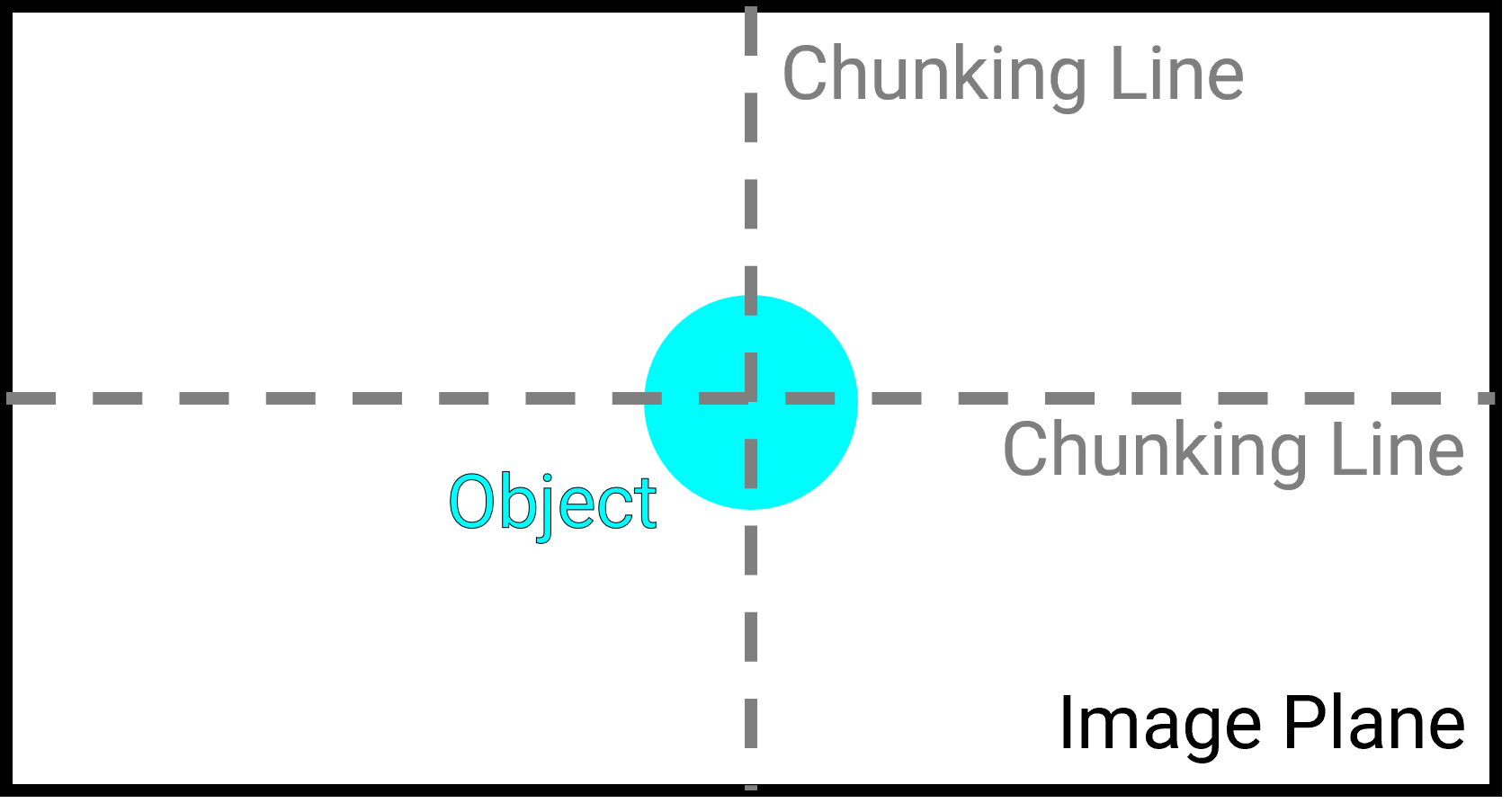}
    \caption{Simplified representation of the scene used for analysis. Notice how the chunking line (both horizontal and vertical) passes through the center of the red object giving rise to a large loss of context for the optical flow network.}
    \label{fig:chukingimg}
\end{figure}

\textbf{Experimental Setup:}
We devised a scenario in Blender$^{\text{\textregistered}}$ where a soccer ball traverses from left to right in the image frame and the camera is perpendicular to the direction of motion while being stationary. We aim to predict the optical flow of this dynamic object within the scene. To ensure detailed analysis, we focused on predicting the optical flow of the ball when it reached the center of the frame (See Fig. \textcolor{red}{S}\ref{fig:chukingimg}). This selection was motivated by the induced loss of spatial information along chunked edges, which is particularly prominent at the center where all inner chunked edges converge. Here, chunked corners lose information from both vertical and horizontal edges.

To examine how this informational loss varies with object size or proximity, we conducted experiments where we increased the distance of the ball from the camera (equivalent to reducing the ball's physical or projected pixel size) and rendered frames capturing its left-to-right movement at the center. To maintain consistency with the paper, each image was rendered at a resolution of $480 \times 352$px. We compared this with chunking methods both with and without overlap. The base chunk resolution remained the same as in the paper ($240 \times 176$), while overlapping pixels were increased by 16, 32, and 64, resulting in final resolutions with overlap of $256 \times 192$px., $272 \times 208$px., and $304 \times 240$px., respectively.

\begin{figure*}[t!]
    \centering
    \includegraphics[width=0.92\textwidth]{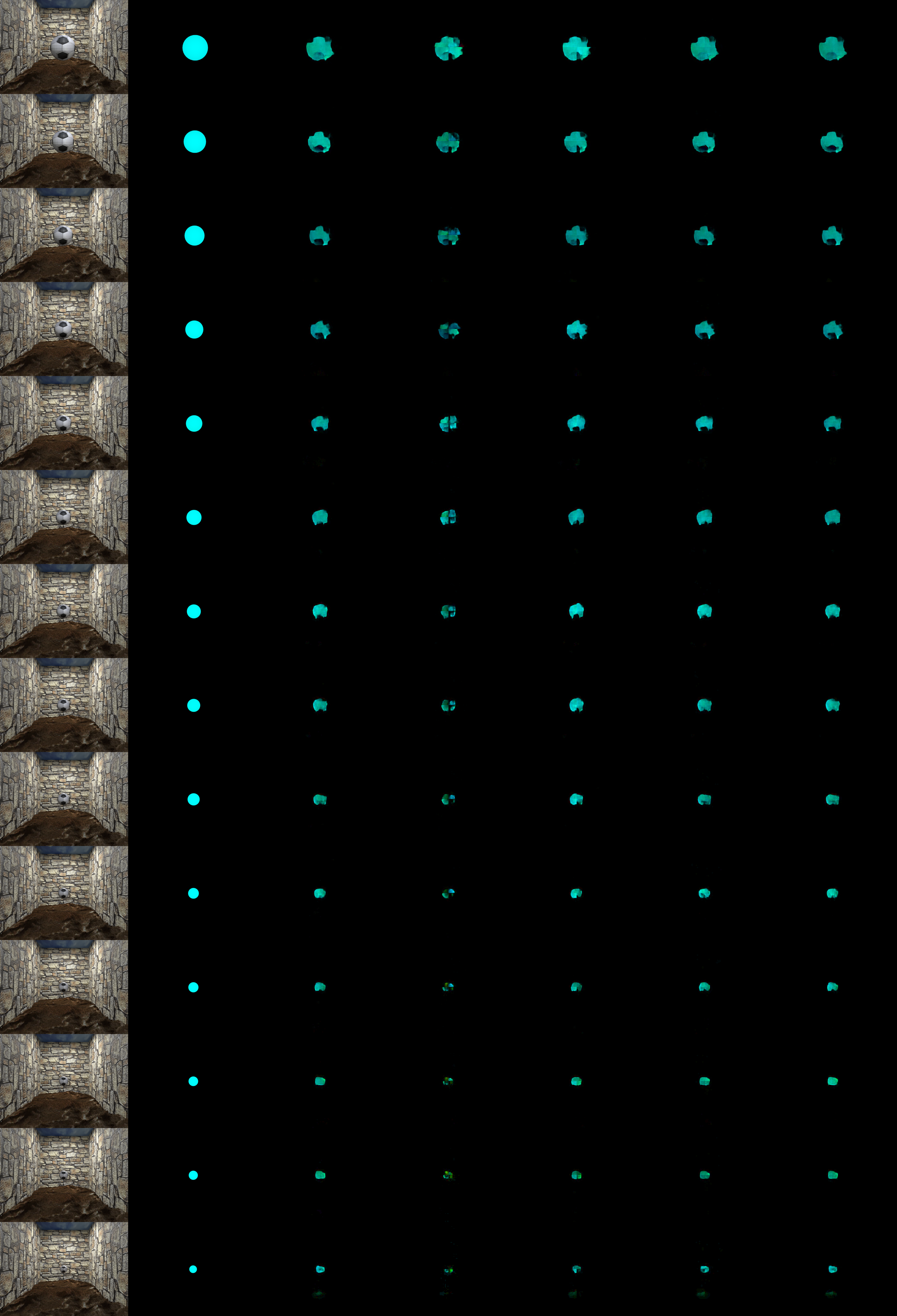}
    \caption{Effects of lack of spatial context in optical flow estimation. Left to right columnwise: Scene image with varying ball sizes, ground truth optical flow, optical flow prediction with full resolution image, prediction with chunking, prediction with overlapping chunking of 16, 32, 64 px. respectively.}
    \label{fig:ball_iou}
\end{figure*}

\textbf{Qualitative Analysis:}
Fig. \textcolor{red}{S}\ref{fig:ball_iou} provides a qualitative analysis of our experiment. Each row in the image represents a different experiment with the ball at varying distances from the camera. Ground truth optical flow was obtained using Blender$^{\text{\textregistered}}$'s cycles rendering. Moving from left to right of the ground truth column, subsequent columns display predictions of \textit{EdgeFlowNet} at different resolutions: full resolution without chunking, with chunking but no overlap, and with chunking and overlap of 16, 32 and 64 pixels, respectively.

Qualitatively, it is evident that chunking decreases the performance of flow prediction as the ball size reduces in the image plane. Conversely, chunking with overlap yields comparable performance to full resolution prediction. Increasing the overlap amount augments the spatial context for the neural network to reason, resulting in better results. It is logical to extend this reasoning to suggest that the required amount of overlapping pixels may vary based on the object size and shape at the center. In our analysis, we opted for a symmetrical shape to simplify the process, allowing for the same amount of overlap in each chunk.

\begin{figure}[t!]
    \centering
    \includegraphics[width=0.5\textwidth]{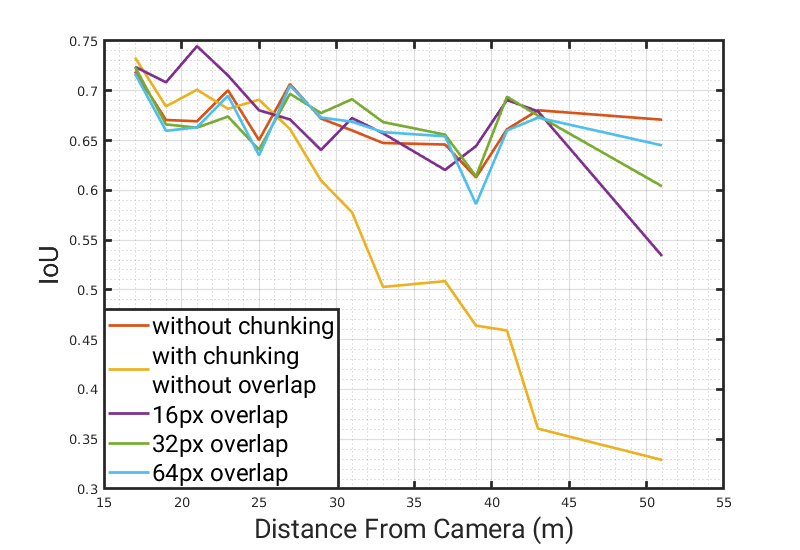}
    \caption{IoU variation in identification of dynamic obstacle (soccer ball) as described in the Fig. \textcolor{red}{S}\ref{fig:ball_iou} without chunking, with chunking and chunking with varying overlapping degrees. Observe how chunking without overlap drops the performance due to lack of spatial context and with overlap the performance is almost on par with that of a full resolution.}
    \label{fig:iou_varition_image}
\end{figure}

\textbf{Quantitative Analysis:}
To provide a more thorough validation, we conduct a quantitative analysis by numerically comparing the predictions with the ground truth (Fig. \textcolor{red}{S}\ref{fig:iou_varition_image}). Following the strategy outlined in the dynamic obstacle avoidance section ($\S$\textcolor{red}{III-C} of the paper), we identify high-magnitude optical flow regions and generate binary masks for ground truth and each type of optical flow prediction discussed earlier. These masks are then compared using the Intersection over Union (IoU) metric.

Fig. \textcolor{red}{S}\ref{fig:ball_iou} depicts the qualitative impact of spatial context loss due to chunking. Clearly, the IoU of the chunking method decreases as the object size reduces in the image plane. As expected, quantitative studies also reveal that overlapping with chunking performs reasonably well and is comparable to full resolution prediction, albeit with a decrease in FPS. Therefore, depending on the task requirements, one can optimize between speed and accuracy by adjusting the amount of overlap.

\section{Architecture Comparison Between \textit{EdgeFlowNet} and NanoFlowNet\cite{bouwmeester2023nanoflownet}}

As noted in the paper, NanoFlowNet\cite{bouwmeester2023nanoflownet} closely parallels our work in terms of utilizing dense optical flow on edge devices and leveraging these predictions for autonomous navigation of small aerial vehicles. Here, we outline the key similarities and differences between our approach and NanoFlowNet:

\subsection{Similarities}
\begin{itemize}
\item \textbf{SWAP Constraints:} Both \textit{EdgeFlowNet} and NanoFlowNet adhere to similar Size, Weight, Area, and Power (SWAP) constraints. NanoFlowNet performs experiments on the off-the-shelf Bitcraze Crazyflie with a GAP8 processor for onboard inferences, while \textit{EdgeFlowNet} utilizes a custom drone with Google's Coral EdgeTPU for onboard inferences.
\item \textbf{Pure CNN Models:} Both architectures are built using convolutional layers. NanoFlowNet primarily employs separable convolution blocks, while \textit{EdgeFlowNet} uses standard convolutional blocks. This choice of pure convolutional blocks is advantageous due to hardware implementation constraints, with convolutional architectures typically offering a good balance between inference speed and accuracy.
\item \textbf{Autonomy Applications:} Both NanoFlowNet and \textit{EdgeFlowNet} target their approaches for deploying autonomy solutions like obstacle avoidance using optical flow on tiny drones.
\end{itemize}

\subsection{Dissimilarities}
\begin{itemize}
\item \textbf{EdgeTPU Compatibility:} Although NanoFlowNet is based on purely CNNs, the combined network graph generated by it is not EdgeTPU compatible. This precludes the possibility of running their network on our EdgeTPU and comparing them directly.
\item \textbf{Resolution, Speed, and Accuracy:} \textit{EdgeFlowNet} can operate at a much higher resolution of $480 \times 352$ while achieving almost real-time speeds (23 FPS) onboard the drone. In comparison, NanoFlowNet's smaller network runs at a resolution of $112 \times 160$ with 10 FPS. Additionally, \textit{EdgeFlowNet} demonstrates higher accuracy with a 6.31px Endpoint Error (EPE) without chunking on MPI-Sintel final, compared to NanoFlowNet's 7.98px EPE.
\end{itemize}

\subsection{Architecture Comparison}

Architecturally, NanoFlowNet and \textit{EdgeFlowNet} diverge significantly from each other. NanoFlowNet draws inspiration from BiSeNet \cite{yu2018bisenet} and STDC-Seg \cite{fan2021rethinking} architectures, while our method is primarily influenced by PRGFlow \cite{sanket2021prgflow}, EVPropNet \cite{sanket2021evpropnet}, and NudgeSeg \cite{singh2021nudgeseg}. Table \textcolor{red}{S}\ref{tab:nanoflownet_arch_diffs} summarizes the main architectural differences between \textit{EdgeFlowNet} and NanoFlowNet. Here $l_1$ and $l_2$ indicate the vector norms of error between ground truth optical flow and predicted optical flow.

\begin{table}[]
\centering
\begin{center}
\caption{Architecture differences between NanoFlowNet and \textit{EdgeFlowNet}.}
\resizebox{\columnwidth}{!}{
\label{tab:nanoflownet_arch_diffs}
\begin{tabular}{ccc}
\toprule
\textbf{Technique} & \textbf{NanoFlowNet} & \textbf{EdgeFlowNet} \\ \hline
\textbf{Backbone} & STDC & ResNet \\ \hline
\textbf{Training Loss} & $l_2 +$ & $l_1 +$  \\
& detail edge loss & self-supervised uncertainty \\ \hline
\textbf{Conceptual} & Low level and & Incremental \\
& High level feature fusion &  multiscale\\ \hline
\textbf{Flow } & Bilinear  & Learnable Convolution \\
\textbf{Upsampling} & interpolation &  Transpose blocks \\ \hline
\textbf{EdgeTPU } & Not & Designed with \\
\textbf{Compatibility} & compatible & EdgeTPU constraints \\ \hline
\textbf{Parameters} & 0.17M & 2M \\ \hline
\textbf{Inference Speed (GPU) FPS} & 48.9 & 98.2 \\ \hline
\textbf{Inference Speed (Edge) FPS} & 5.6 (GAP8) & 93.6 (EdgeTPU) \\
\bottomrule\\[-16pt]
\end{tabular}
}
\end{center}
\end{table}

\section{Network ablation study for EdgeTPU}
\begin{figure}[t!]
    \centering
    \includegraphics[width=1\linewidth]{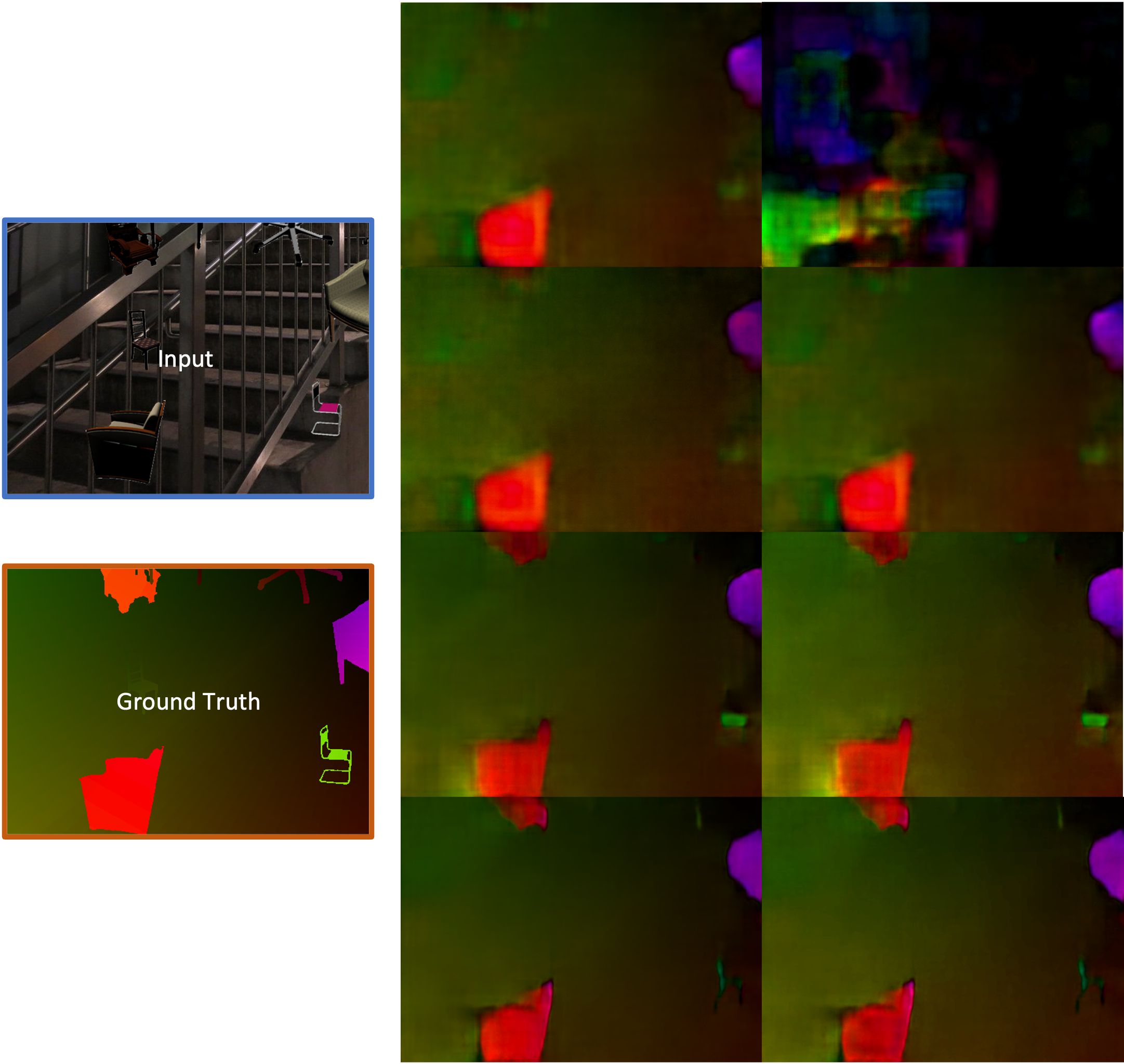}
    \caption{Second column is GPU, third column is EdgeTPU. Last two columns, from top to bottom, network predictions for $l_1$, $l_1 + 50$, $l_1$ Multiscale, $l_1$ Multiscale + Uncertainty.}
    \label{fig:network_analysis}
\end{figure}

In this section, we briefly mention different network architectures and losses that we experimented with to check for EdgeTPU compatibility. Table \textcolor{red}{S}\ref{tab:ablation_study} highlights several insights about EdgeTPU-compatible architectures. Each architecture is trained on FlyingChairs2 dataset for 400 epochs using different losses with a learning rate of $10^{-4}$ and a batch size of 32. The $l_1$ and $l_1 + 50$ (level-shifted optical flow to make it fit in \texttt{UINT8} range) methods utilize pure convolution and convolution transpose networks without multiscale connections as described in the paper. While the $l_1$ method performs comparably well on GPU, it exhibits adverse results on the EdgeTPU platform. Conversely, $l_1 + 50$ produces outputs modeled in the completely positive domain and performs well on both GPU and EdgeTPU. This underscores how the limited unsigned 8-bit integer range on EdgeTPU affects regression unless explicitly modeled in the network. This is also evident in the first two rows of Fig. \textcolor{red}{S}\ref{fig:network_analysis}.

The $l_1$ Multiscale and $l_1$ Multiscale + Uncertainty methods predict incremental flow without shifted domains, highlighting how this approach ensures that the network learns to predict differences between scales within the constrained range of unsigned 8-bit numerical values (0 to 255) after EdgeTPU conversion, thereby preventing overflow concerns. The qualitative performance of the multiscale architectures can be seen in the last two rows of the Fig. \textcolor{red}{S}\ref{fig:network_analysis}. 

\begin{table}[t!]
\centering
\caption{EdgeTPU compatible Network ablation study on FlyingChairs2 dataset.}
\resizebox{\columnwidth}{!}
{
\label{tab:ablation_study}
\begin{tabular}{lcccccc}
\toprule
Loss & Num.   & Num.   &  \multicolumn{2}{c}{GPU}  &    \multicolumn{2}{c}{EdgeTPU}     \\
\cline{4-7}
     & Epochs & Params & FPS$\uparrow$ & EPE$\downarrow$ &    FPS$\uparrow$      &    EPE$\downarrow$      \\
  \hline
$l_1$                            & 400 & 2M   & \textbf{158}              & 4.31 & \textbf{64}  & 20.1     \\
$l_1+50$                         & 400 & 2M   & 117            & 4.35  & 14.4 & 5.31                     \\
$l_1$ MultiScale                 & 400 & 2M   & 99.8           & 2.85  & 21.3 & 3.60  \\
$l_1$ MultiScale + Uncertainity  & 400 & 2M   & 98.4           & \textbf{2.76}  & 22.8 & \textbf{3.19}                     \\
 \bottomrule
\end{tabular}}

\end{table}

\section{Impact of resolution on the inference speed}
Here, we perform an experiment to see how input/output loading and caching affects the network inference time for a single convolution layer. To this end, we feed images of different resolutions into a quantized convolutional layer on the Google Coral EdgeTPU. The input resolutions are varied as powers of two since this is the most cache friendly number and the inference time is tabulated. The inference time includes the time for loading/unloading images into EdgeTPU cache and the time for computation. Table \textcolor{red}{S}\ref{tab:network_inference_speed} demonstrates the speeds for chunking approach. From the table, we observe that inference speed increases as the number of chunks increases and the resolution decreases. However, there is a drop in speed for the input size $4096 \times 16 \times 16 \times 6$. This behavior aligns with the concepts of cache hits and cache misses: as long as the EdgeTPU can efficiently load the input data without reloading from the main memory, it delivers optimized performance. When it reaches a point where it cannot load enough data to fit in the cache line, it has to unload and reload existing data, which consumes clock cycles and increases latency.

\begin{table}[t!]
\centering
\caption{Impact of resolution on the inference speed}
\resizebox{0.7\columnwidth}{!}
{
\label{tab:network_inference_speed}
\begin{tabular}{lcc}
\toprule
Input Size & Chunking FPS$\uparrow$    \\
  \hline
$1 \times 1024 \times 1024 \times 6$ & 7.68    \\
$4 \times 512 \times 512 \times 6$   & 13.33   \\
$16 \times 256 \times 256 \times 6$   & 55.74  \\
$64 \times 128 \times 128 \times 6$  & 153.98  \\
$256 \times 64 \times 64 \times 6$   & 339.11  \\
$1024 \times 32 \times 32 \times 6$  & 444.27  \\
$4096\times 16 \times 16 \times 6$    & 321.79 \\
 \bottomrule
\end{tabular}}

\end{table}

\section{Chunking Variation across GPUs}

Table \textcolor{red}{S}\ref{tab:network_inference_speed2} presents the chunking speeds, measured in Frames Per Second (FPS), across three different GPU architectures: the NVIDIA 3060Ti, NVIDIA 4070, and NVIDIA 4090. The input sizes represent varying levels of chunking, with larger chunks corresponding to smaller batch sizes. This table highlights the differences in chunking speeds across different GPU architectures. We observe speedups across all GPUs when the resolution decreases but to different extents. Notably, we observe smaller speedups with larger memory GPUs since they are optimized for larger spatial resolutions rather than smaller ones. Specifically, the overhead to load/unload data to/from the GPU is higher with the larger VRAM size. On the contrary, EdgeTPU is optimized for smaller spatial resolutions, giving us massive speedups when chunking is used.  Notably, the larger VRAM NVIDIA 4070 and NVIDIA 4090 GPUs show a slower inference speed at $256 \times 16 \times 16 \times 6$  as compared to $64 \times 16 \times 16 \times 6$. We believe this is due to the large overhead times to catching and load/unload data times to/from the GPUs which dominate the time as compared to compute time.

\begin{table}[t!]
\centering
\caption{EdgeFlowNet chunking speeds on different GPUs}
\resizebox{\columnwidth}{!}
{
\label{tab:network_inference_speed2}
\begin{tabular}{lcccc}
\toprule
Input Size & EdgeTPU FPS$\uparrow$ & 3060Ti FPS$\uparrow$    & 4070 FPS$\uparrow$ & 4090 FPS$\uparrow$\\
  \hline
$1 \times 480 \times 352 \times 6$ & 22.8 & 98.3 & 160.1   & \textbf{194.6} \\
$4 \times 240 \times 176 \times 6$ & 93.4 & 98.2  & 162.2  & \textbf{176.4}\\
$16 \times 128 \times 96 \times 6$ & \textbf{283.7} & 170.6  & 216.5 & 223.1\\
$64 \times 64 \times 48 \times 6$  & \textbf{794.5} & 217.4 & 238.2 & 280.0\\
$256 \times 16 \times 16 \times 6$ & \textbf{1959.3} & 240.9 & 231.5 & 256.5\\
 \bottomrule
\end{tabular}}
\end{table}


\section{Chunking with Different Architectures}

Table \textcolor{red}{S}\ref{tab:edgetpu_resolutions_comparison2} presents an analysis of FPS performance across different network architectures when utilizing EdgeTPU. We chose three commonly used architectures (that are EdgeTPU compatible) and performed analysis similar to \cite{prgflow}. These architectures are VanillaNet \cite{vanillanet}, SqueezeNet \cite{squeezenet} and MobileNet \cite{mobilenet}. We utilize similar architecture configurations as \cite{prgflow} and add mirrored decoder blocks with convolution operations replaced by transposed convolution operations. We keep the overall number of parameters almost the same across architectures.

The results highlight the variation in FPS across the models. They also indicate that chunking leads to improved performance and is network architecture agnostic. The extent of performance gains does vary depending on the model size and design. Chunking process can be hence treated as a generalized blueprint approach for speeding up networks when running on the Google Coral EdgeTPU.

\begin{table}[h!]
\centering
\caption{EdgeTPU Chunking on different networks and resolutions}
\label{tab:edgetpu_resolutions_comparison2}
\resizebox{\columnwidth}{!}
{
\begin{tabular}{@{}lcccc@{}}
    \toprule
    {Input Resolution} & \textit{EdgeFlowNet (Ours)} & {VanillaNet \cite{vanillanet}} & {SqueezeNet \cite{squeezenet}} & {MobileNet \cite{mobilenet}} \\ 
    \midrule
    \textbf{$1 \times 480 \times 352 \times 6$}  & 22.8  & 12.3  & 6.7   & 30.1   \\
    \textbf{$4 \times 240 \times 176 \times 6$}  & 93.4   & 134.1 & 26.3   & 150.2 \\
    \textbf{$16 \times 128 \times 96 \times 6$}  & 283.7     & 387.2     & 199.0     & 487.8     \\
    \textbf{$64 \times 64 \times 48 \times 6$}   & 794.5     & 1029.5     & 705.0     & 1248.6     \\
    \textbf{$256 \times 16 \times 16 \times 6$}  & \textbf{1959.3}     & \textbf{2993.0}     & \textbf{1857.2}     & \textbf{3102.2}     \\
    \bottomrule
\end{tabular}}
\end{table}

\end{document}